\documentclass{article}

\usepackage[preprint]{corl_2023} 
\usepackage{enumitem}
\usepackage{graphicx}
\usepackage{subfig}
\usepackage{booktabs} 
\usepackage{multirow}
\usepackage{wrapfig}
\setlist[itemize,enumerate]{topsep=1pt,itemsep=2pt}

\usepackage{titlesec}
\titlespacing*{\subsection}{0pt}{0.1\baselineskip}{0.05\baselineskip}
\titlespacing*{\section}{0pt}{0.2\baselineskip}{0.1\baselineskip}

\title{Decomposing the Generalization Gap in\\Imitation Learning for Visual Robotic Manipulation}

\author{
  Annie Xie$^{*1}$ \quad Lisa Lee$^{*2}$ \quad Ted Xiao$^2$ \quad Chelsea Finn$^{1,2}$\\
  $^1$Stanford University \qquad $^2$Google DeepMind
}

\begin{document}
\maketitle


\begin{abstract}
    What makes generalization hard for imitation learning in visual robotic manipulation? This question is difficult to approach at face value, but the environment from the perspective of a robot can often be decomposed into enumerable \emph{factors of variation}, such as the lighting conditions or the placement of the camera. Empirically, generalization to some of these factors have presented a greater obstacle than others, but existing work sheds little light on precisely how much each factor contributes to the generalization gap. Towards an answer to this question, we study imitation learning policies in simulation and on a real robot language-conditioned manipulation task to quantify the difficulty of generalization to different (sets of) factors. We also design a new simulated benchmark of 19 tasks with 11 factors of variation to facilitate more controlled evaluations of generalization. From our study, we determine an ordering of factors based on generalization difficulty, that is consistent across simulation and our real robot setup.\footnote{Videos \& code are available at: \url{https://sites.google.com/view/generalization-gap}}
\end{abstract}



\section{Introduction}

\begin{wrapfigure}{r}{0.4\linewidth}
    \centering
    \vspace{-0.5cm}
    \includegraphics[width=\linewidth]{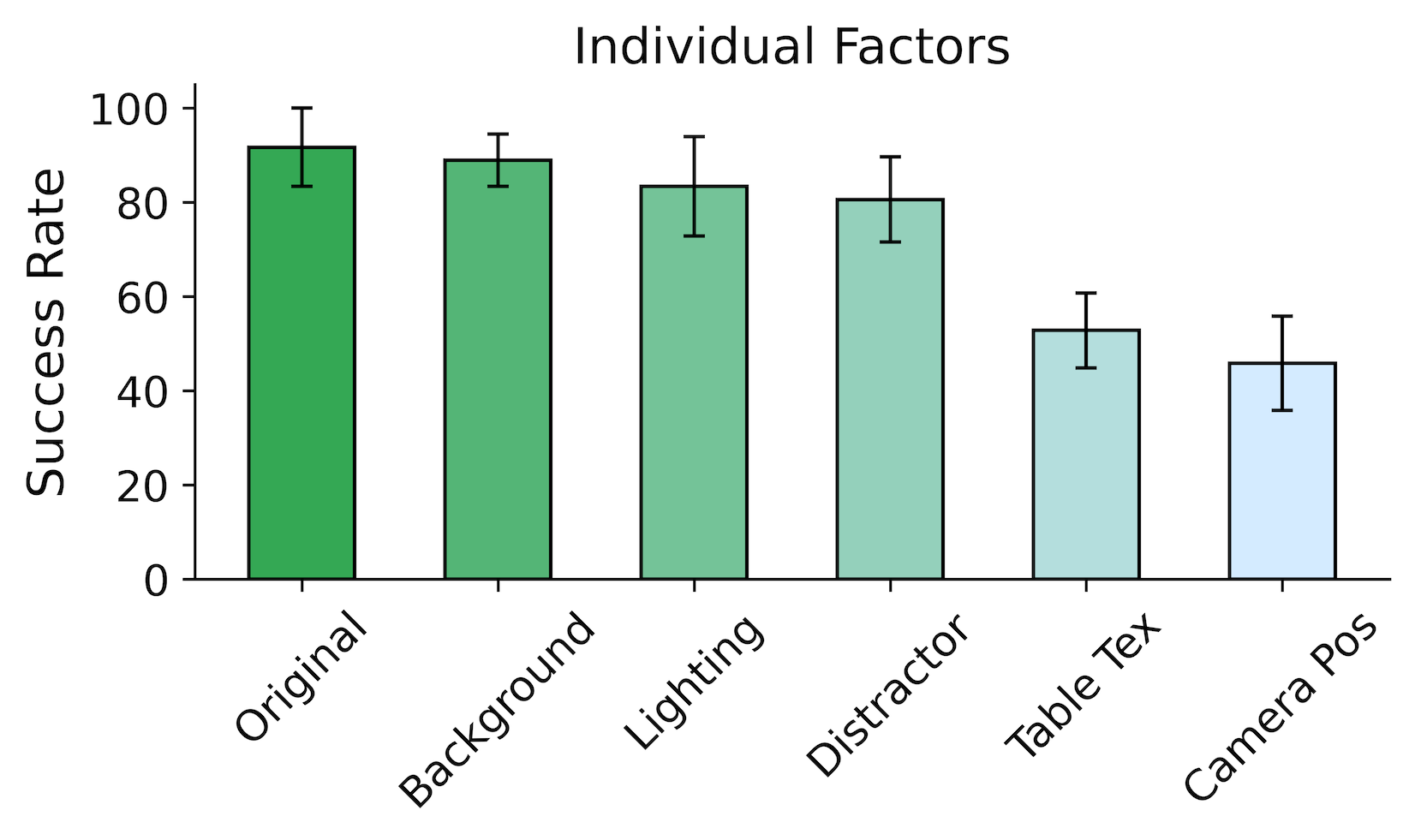}
    \vspace{-0.6cm}
    \caption{\small Success rates on different environment shifts. New camera positions are the hardest to generalize to while new backgrounds are the easiest.}
    \vspace{-0.2cm}
    \label{fig:real_single_factor_bar}
\end{wrapfigure}
Robotic policies often fail to generalize to new environments, even after training on similar contexts and conditions. In robotic manipulation, data augmentation techniques~\citep{laskin2020reinforcement,yarats2021mastering,hansen2021generalization,young2021visual,graf2022learning} and representations pre-trained on large datasets~\citep{yen2020learning,radford2021learning,khandelwal2022simple,shridhar2022cliport,shah2021rrl,parisi2022unsurprising,nair2022r3m} improve performance but a gap still remains. Simultaneously, there has also been a focus on the collection and curation of reusable robotic datasets~\citep{sharma2018multiple,mandlekar2018roboturk,mandlekar2019scaling,dasari2019robonet,ebert2021bridge}, but there lacks a consensus on how much more data, and what \emph{kind} of data, is needed for good generalization. These efforts could be made significantly more productive with a better understanding of which dimensions existing models struggle with. Hence, this work seeks to answer the question: \emph{What are the factors that contribute most to the difficulty of generalization to new environments in vision-based robotic manipulation?}

To approach this question, we characterize environmental variations as a combination of independent factors, namely the background, lighting condition, distractor objects, table texture, object texture, table position, and camera position. This decomposition allows us to quantify how much each factor contributes to the generalization gap, which we analyze in the imitation learning setting (see Fig.~\ref{fig:real_single_factor_bar} for a summary of our real robot evaluations). While vision models are robust to many of these factors already~\citep{hendrycks2019benchmarking,hendrycks2021many,geirhos2021partial}, robotic policies are considerably less mature, due to the smaller and less varied datasets they train on. In robot learning, data collection is largely an \emph{active} process, in which robotics researchers design and control the environment the robot interacts with. As a result, naturally occurring variations, such as different backgrounds, are missing in many robotics datasets. Finally, robotics tasks require dynamic, multi-step decisions, unlike computer vision tasks such as image classification. These differences motivate our formal study of these environment factors in the context of robotic manipulation.

In our study, we evaluate a real robot manipulator on over $20$ test scenarios featuring new lighting conditions, distractor objects, backgrounds, table textures, and camera positions. We also design a suite of $19$ simulated tasks, equipped with $11$ customizable environment factors, which we call \emph{Factor World}, to supplement our study. With over $100$ configurations for each factor, \emph{Factor World} is a rich benchmark for evaluating generalization, which we hope will facilitate more fine-grained evaluations of new models, reveal potential areas of improvement, and inform future model design. Our study reveals the following insights:

\begin{itemize}[leftmargin=0.5cm]
    \item \emph{Most pairs of factors do not have a compounding effect on generalization performance.
    } For example, generalizing to the combination of new table textures and new distractor objects is no harder than new table textures alone, which is the harder of two factors to generalize to. This result implies that we can study and address environment factors individually.

    \item \emph{Random crop augmentation improves generalization even along non-spatial factors.} We find that random crop augmentation is a lightweight way to improve generalization to spatial factors such as camera positions, but also to non-spatial factors such as distractor objects and table textures.

    \item \emph{Visual diversity from out-of-domain data dramatically improves generalization.} In our experiments, we find that training on data from other tasks and domains like opening a fridge and operating a cereal dispenser can improve performance on picking an object from a tabletop. 
\end{itemize}

\section{Related Work}
\label{sec:related_work}

Prior efforts to address robotic generalization include diverse datasets, pretrained representations, and data augmentation, which we discuss below. 

\textbf{Datasets and benchmarks.} Existing robotics datasets exhibit rich diversity along multiple dimensions, including objects~\citep{kalashnikov2018scalable,dasari2019robonet,ebert2021bridge,brohan2022rt}, domains~\citep{dasari2019robonet,young2021visual,ebert2021bridge}, and tasks~\citep{sharma2018multiple,mandlekar2018roboturk,mandlekar2019scaling}. However, collecting high-quality and diverse data \emph{at scale} is still an unsolved challenge, which motivates the question of how new data should be collected given its current cost. The goal of this study is to systematically understand the challenges of generalization to new objects and domains\footnote{We define an environment to be the combination of the domain and its objects.} and, through our findings, inform future data collection strategies.
Simulation can also be a useful tool for understanding the scaling relationship between data diversity and policy performance, as diversity in simulation comes at a much lower cost~\citep{tan2018sim,peng2018sim,chebotar2019closing,james2019sim}. Many existing benchmarks aim to study exactly this~\citep{yu2020meta,cobbe2020leveraging,stone2021distracting,xing2021kitchenshift}; these benchmarks evaluate the generalization performance of control policies to new tasks~\citep{yu2020meta,cobbe2020leveraging} and environments~\citep{stone2021distracting,xing2021kitchenshift}.~\emph{KitchenShift}~\citep{xing2021kitchenshift} is the most related to our contribution~\emph{Factor World}, benchmarking robustness to shifts like lighting, camera view, and texture. However,~\emph{Factor World} contains a more complete set of factors ($11$ versus $7$ in~\emph{KitchenShift}) with many more configurations of each factor (over $100$ versus fewer than $10$ in~\emph{KitchenShift}). 

\textbf{Pretrained representations and data augmentation.} 
Because robotics datasets are generally collected in fewer and less varied environments, prior work has leveraged the diversity found in large-scale datasets from other domains like static images from ImageNet~\citep{russakovsky2015imagenet}, videos of humans from Ego4D~\citep{grauman2022ego4d}, and natural language~\citep{yen2020learning,shridhar2022cliport,khandelwal2022simple,shah2021rrl,nair2022r3m}. While these datasets do not feature a single robot, pretraining representations on them can lead to highly efficient robotic policies with only a few episodes of robot data~\citep{khandelwal2022simple,nair2022r3m,ma2022vip}. A simpler yet effective way to improve generalization is to apply image data augmentation techniques typically used in computer vision tasks~\citep{chen2020simple}. Augmentations like random shifts, color jitter, and rotations have been found beneficial in many image-based robotic settings~\citep{laskin2020reinforcement,kostrikov2020image,yarats2021mastering,hansen2021stabilizing,hansen2021generalization,young2021visual}. While pretrained representations and data augmentations have demonstrated impressive empirical gains in many settings, we seek to understand when and why they help, through our factor decomposition of robotic environments. 

\textbf{Generalization in RL.} Many generalization challenges found in the RL setting are shared by the imitation learning setting and vice versa. Common approaches in RL include data augmentation~\citep{laskin2020reinforcement,kostrikov2020image,hansen2021stabilizing}, domain randomization~\citep{rajeswaran2016epopt,pinto2017robust,tan2018sim}, and modifications to the network architecture~\citep{cobbe2020leveraging,raileanu2021decoupling,cetin2022stabilizing}. We refer readers to~\citeauthor{kirk2021survey} for a more thorough survey of challenges and solutions. Notably,~\citeauthor{packer2018assessing} also conducted a study to evaluate the generalization performance of RL policies to new physical parameters, such as the agent's mass, in low-dimensional tasks~\citep{todorov2012mujoco}. Our work also considers the effect of physical parameters, such as the table position, but because our tasks are solved from image-based observations, these changes to the environment are observed by the agent. We instead evaluate the agent's ability to generalize to these observable changes.

\section{Environment Factors}
\label{sec:factors}

Several prior works have studied the robustness of robotic policies to different environmental shifts, such as harsher lighting, new backgrounds, and new distractor objects~\citep{julian2020never,xing2021kitchenshift,brohan2022rt,zhou2022train}. Many interesting observations have emerged from them, such as how mild lighting changes have little impact on performance~\citep{julian2020never} and how new backgrounds (in their case, new kitchen countertops) have a bigger impact than new distractor objects~\citep{brohan2022rt}. However, these findings are often qualitative or lack specificity. For example, the performance on a new kitchen countertop could be attributed to either the appearance or the height of the new counter. A goal of our study is to formalize these prior observations through systematic evaluations and to extend them with a more comprehensive and fine-grained set of environmental shifts. In the remainder of this section, we describe the environmental factors we evaluate and how we implement them in our study. 

\begin{figure*}
    \centering
    \subfloat[Original\label{fig:envs-original}]{
        \includegraphics[width=0.188\textwidth,trim={0 0 0 0},clip]{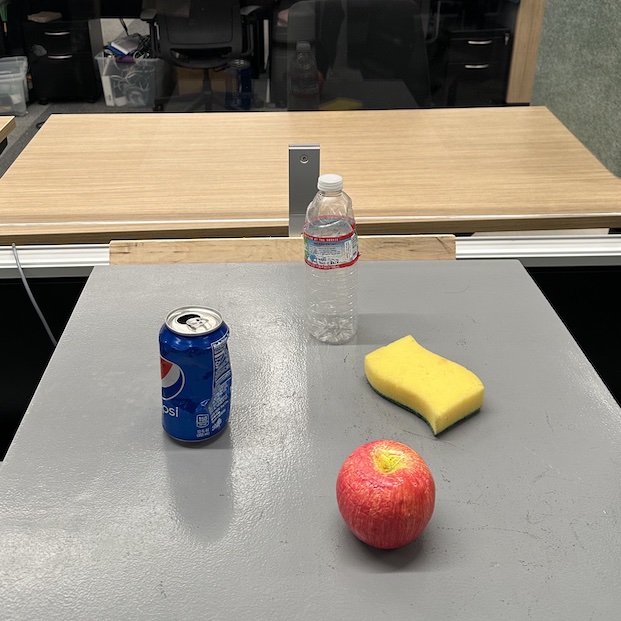}
    }
    \subfloat[Light\label{fig:envs-light}]{
      \includegraphics[width=0.188\textwidth,trim={0 0 0 0},clip]{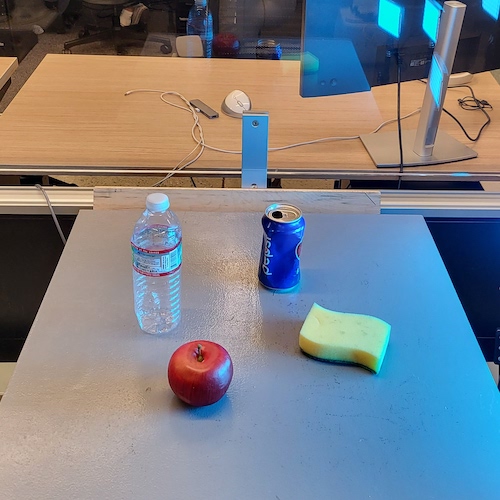}
    }
    \subfloat[Distractors\label{fig:envs-distractor}]{
      \includegraphics[width=0.188\textwidth,trim={0 0 0 0},clip]{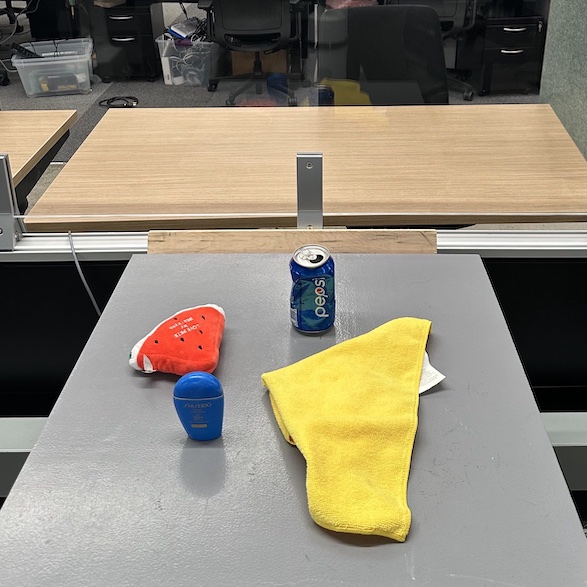}
    }
    \subfloat[Table texture\label{fig:envs-table}]{
      \includegraphics[width=0.188\textwidth,trim={0 0 0 0},clip]{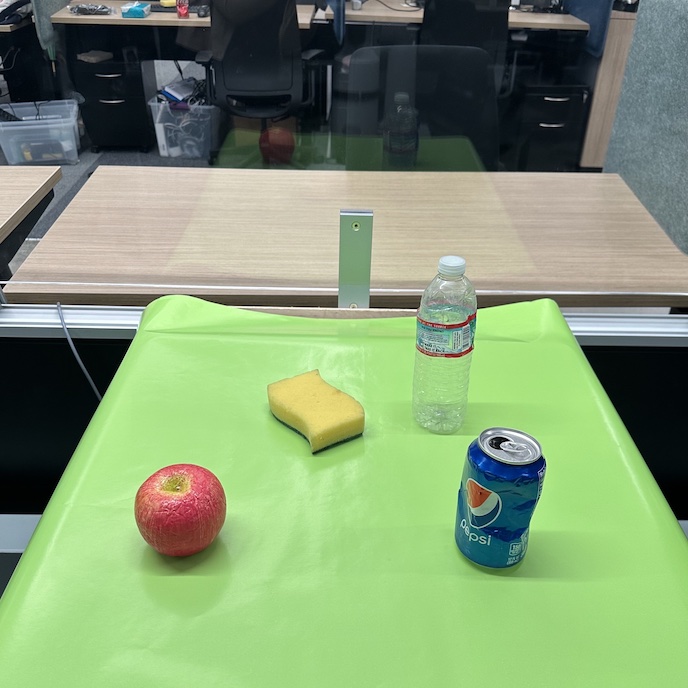}
    }
    \subfloat[Background\label{fig:envs-background}]{
      \includegraphics[width=0.188\textwidth,trim={0 0 0 0},clip]{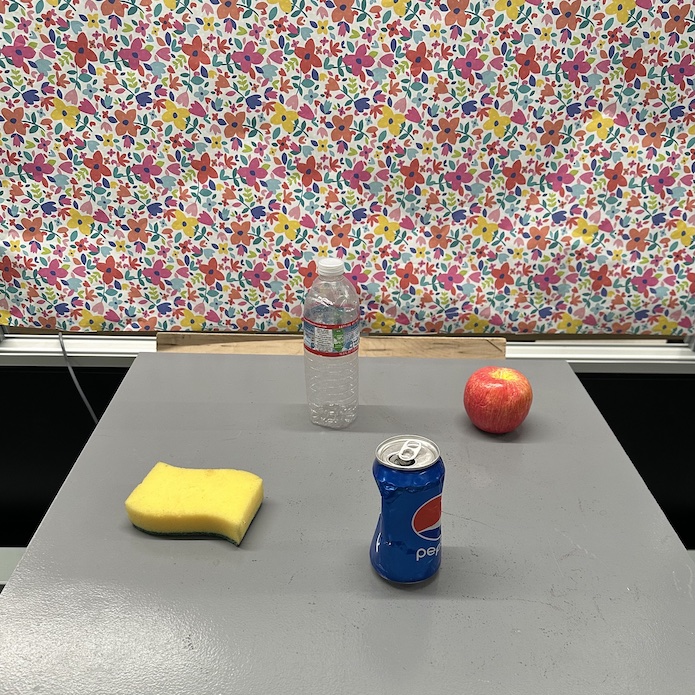}
    }
    \caption{\small Examples of our real robot evaluation environment. We systematically vary different environment factors, including the lighting condition, distractor objects, table texture, background, and camera pose.}
    \label{fig:robot_factors}
    \vspace{-0.2cm}
\end{figure*}
\begin{figure*}
    \centering
    \subfloat[Light setup 1\label{fig:setup_light1}]{
        \includegraphics[width=0.188\textwidth]{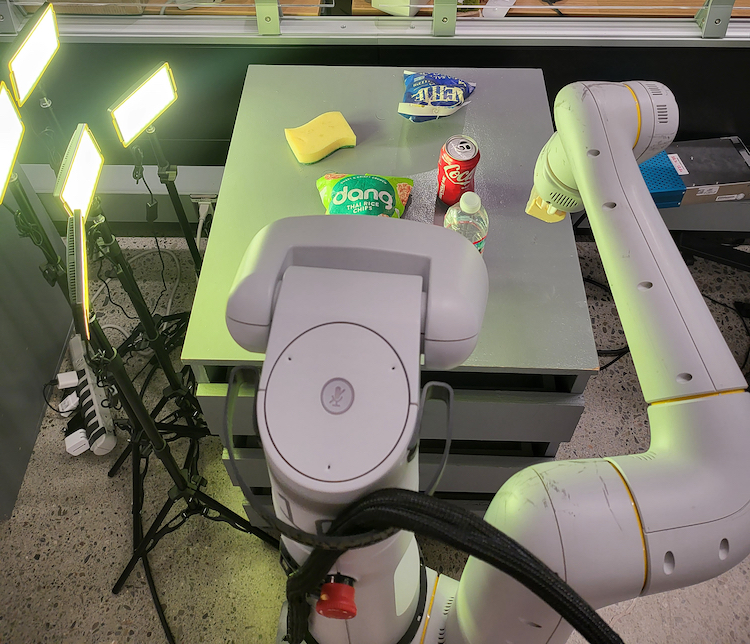}
    }
    \subfloat[Light setup 2\label{fig:setup_light2}]{
        \includegraphics[width=0.188\textwidth]{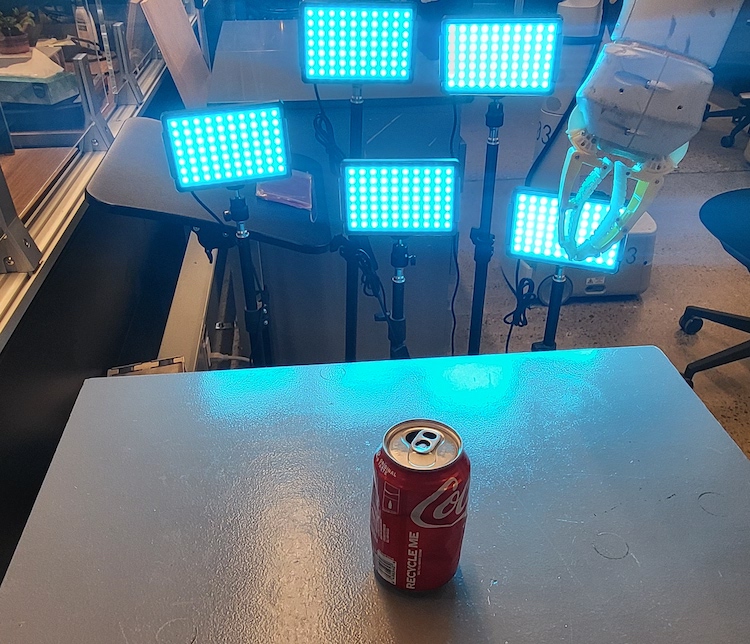}
    }
    \subfloat[Original view\label{fig:camera1}]{
      \includegraphics[width=0.188\textwidth]{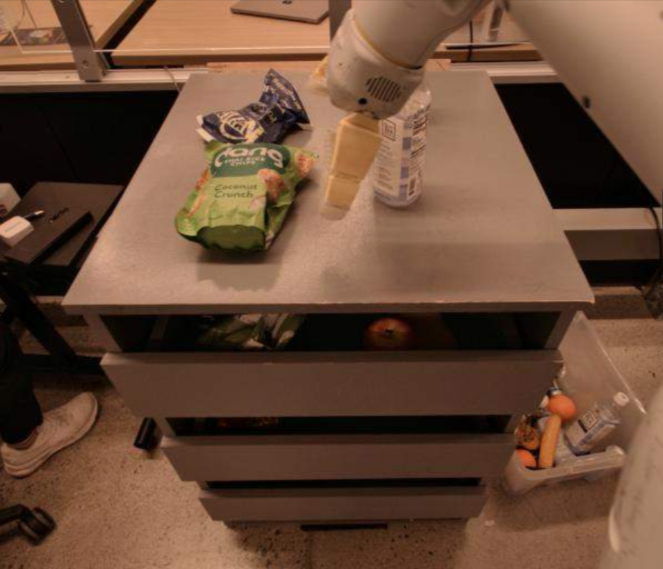}
    }
    \subfloat[Test view 1\label{fig:camera2}]{
      \includegraphics[width=0.188\textwidth]{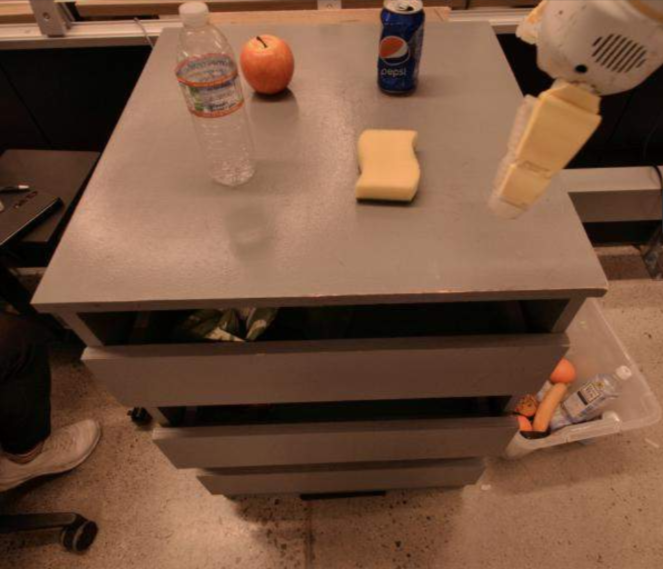}
    }
    \subfloat[Test view 2\label{fig:camera3}]{
      \includegraphics[width=0.188\textwidth]{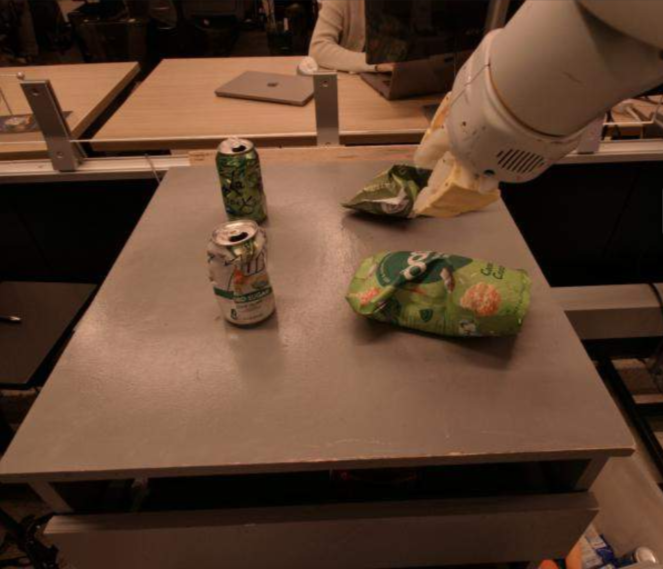}
    }
    \caption{\small (a-b) Our setup to evaluate changes in lighting. (c-e) The original and new camera views.}
    \label{fig:distractors_and_camera}
    \vspace{-0.2cm}
\end{figure*}

\subsection{Real Robot Manipulation}
In our real robot evaluations, we study the following factors: lighting condition, distractor objects, background, table texture, and camera pose. In addition to selecting factors that are specific and controllable, we also take inspiration from prior work, which has studied robustness to many of these shifts~\citep{julian2020never,xing2021kitchenshift,brohan2022rt}, thus signifying their relevance in real-world scenarios.

Our experiments are conducted with mobile manipulators. The robot has a right-side arm with seven DoFs, gripper with two fingers, mobile base, and head with integrated cameras. The environment, visualized in Fig.~\ref{fig:envs-original}, consists of a cabinet top that serves as the robot workspace and an acrylic wall that separates the workspace and office background. To control the lighting condition in our evaluations, we use several bright LED light sources with different colored filters to create colored hues and new shadows (see Fig.~\ref{fig:envs-light}). We introduce new table textures and backgrounds by covering the cabinet top and acrylic wall, respectively, with patterned paper. We also shift the camera pose by changing the robot's head orientation (see Fig.~\ref{fig:distractors_and_camera} for the on-robot perspectives from different camera poses). Due to the practical challenges of studying factors like the table position and height, we reserve them for our simulated experiments.

\subsection{Factor World}
\begin{figure*}[t]
\centering
\subfloat[Pick Place\label{fig:envs-pick_place}]{
  \includegraphics[width=0.188\textwidth,trim={20pt 30pt 50pt 20pt},clip]{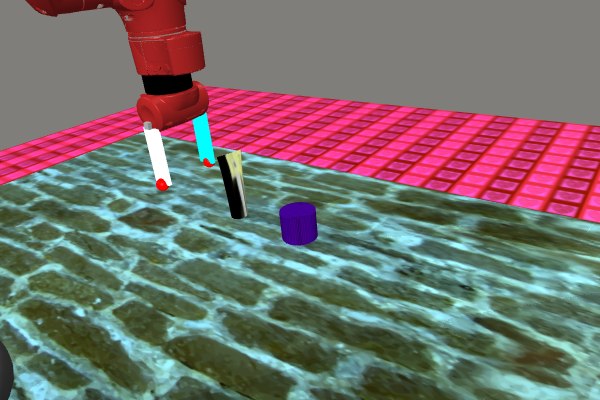}
}
\subfloat[Bin Picking\label{fig:envs-bin_picking}]{
  \includegraphics[width=0.188\textwidth,trim={20pt 30pt 50pt 20pt},clip]{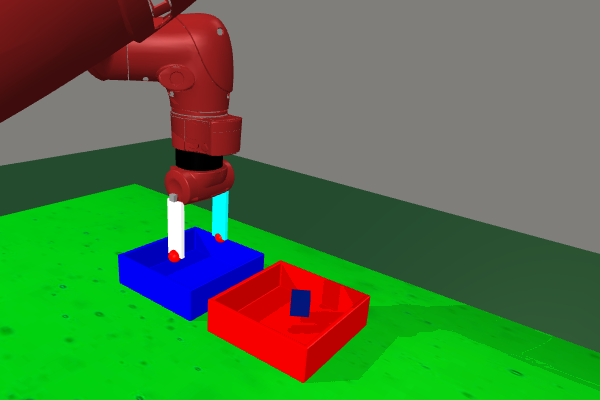}
}
\subfloat[Door {\tiny(Open, Lock)}\label{fig:envs-door}]{
  \includegraphics[width=0.185\textwidth,trim={20pt 30pt 50pt 20pt},clip]{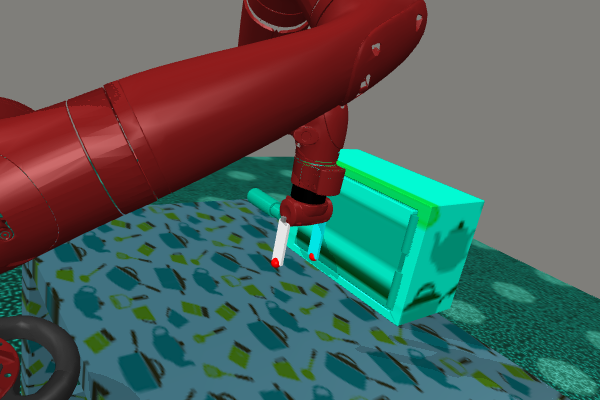}
}
\subfloat[Basketball\label{fig:envs-basketball}]{
  \includegraphics[width=0.188\textwidth,trim={20pt 30pt 50pt 20pt},clip]{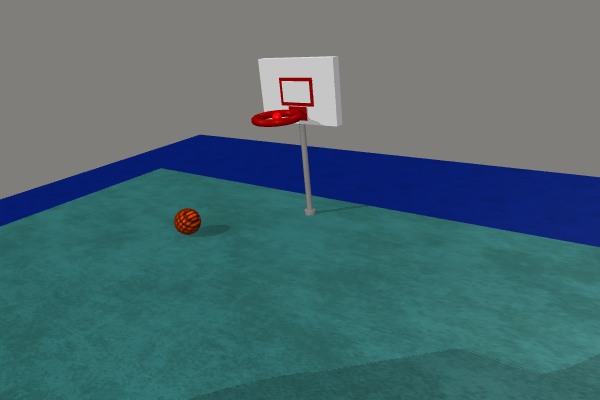}
}
\subfloat[Button {\tiny(Top, Side, Wall)}\label{fig:envs-button}]{
  \includegraphics[width=0.188\textwidth,trim={20pt 30pt 50pt 20pt},clip]{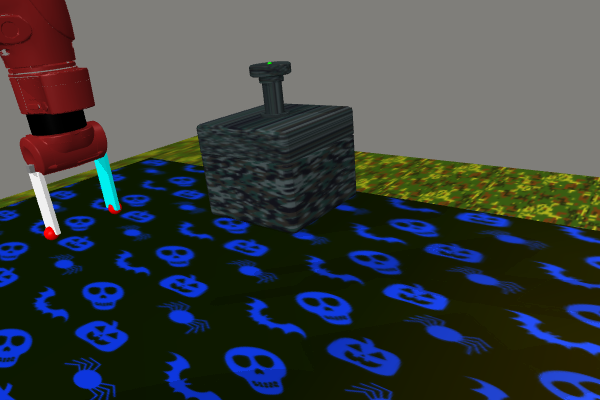}
}
\caption{\small \emph{Factor World}, a suite of 19 visually diverse robotic manipulation tasks. Each task can be configured with multiple factors of variation such as lighting; texture, size, shape, and initial position of objects; texture of background (table, floor); position of the camera and table relative to the robot; and distractor objects.}
\label{fig:sim_envs}
\vspace{-0.2cm}
\end{figure*}

We implement the environmental shifts on top of the \emph{Meta World} benchmark~\citep{yu2020meta}. While \emph{Meta World} is rich in diversity of control behaviors, it lacks diversity in the environment, placing the same table at the same position against the same background. Hence, we implement 11 different factors of variation, visualized in Fig.~\ref{fig:sim_envs} and fully enumerated in Fig.~\ref{fig:factors-of-variation}. These include lighting; texture, size, shape, and initial position of objects; texture of the table and background; the camera pose and table position relative to the robot; the initial arm pose; and distractor objects. In our study, we exclude the object size and shape, as an expert policy that can handle any object is more difficult to design, and the initial arm pose, as this can usually be fixed whereas the same control cannot be exercised over the other factors, which are inherent to the environment.

Textures (table, floor, objects) are sampled from 162 texture images (81 for train, 81 for eval) and continuous RGB values in $[0, 1]^3$. Distractor objects are sampled from 170 object meshes (100 for train, 70 for eval) in Google's Scanned Objects Dataset~\citep{zakka2022scannedobjectsmujoco,downs2022scannedobjects}. For lighting, we sample continuous ambient and diffuse values in $[0.2, 0.8]$. Positions (object, camera, table) are sampled from continuous ranges summarized in Table~\ref{tbl:continuous}. We consider different-sized ranges to control the difficulty of generalization. While fixing the initial position of an object across trials is feasible with a simulator, it is generally difficult to precisely replace an object to its original position in physical setups.  Thus, we randomize the initial position of the object in each episode in the experiments.

\section{Study Design}

We seek to understand how each environment factor described in Sec.~\ref{sec:factors} contributes to the difficulty of generalization. In our pursuit of an answer, we aim to replicate, to the best of our ability, the scenarios that robotics practitioners are likely to encounter in the real world. We therefore start by selecting a set of tasks commonly studied in the robotics literature and the data collection procedure (Sec.~\ref{subsec:tasks}). Then, we describe the algorithms studied and our evaluation protocol (Sec.~\ref{subsec:algorithms}).

\subsection{Control Tasks and Datasets}
\label{subsec:tasks}

\emph{Real robot manipulation.} We study the language-conditioned manipulation problem from~\citeauthor{brohan2022rt}, specifically, focusing on the ``pick'' skill for which we have the most data available. The goal is to pick up the object specified in the language instruction. For example, when given the instruction ``pick pepsi can'', the robot should pick up the pepsi can among the distractor objects from the countertop 
(Fig.~\ref{fig:robot_factors}). We select six objects for our evaluation; all ``pick'' instructions can be found in Fig.~\ref{fig:pick_tasks} in App.~\ref{app:exp_details}. The observation consists of $300 \times 300$ RGB image observations from the last six time-steps and the language instruction, while the action controls movements of the arm ($xyz$-position, roll, pitch, yaw, opening of the gripper) and movements of the base ($xy$-position, yaw). The actions are discretized along each of the $10$ dimensions into $256$ uniform bins. The real robot manipulation dataset consists of over $115$K human-collected demonstrations, collected across $13$ skills, with over $100$ objects, three tables, and three locations. The dataset is collected with a fixed camera orientation but randomized initial base position in each episode.

\emph{Factor World.} While \emph{Factor World} consists of 19 robotic manipulation tasks, we focus our study on three commonly studied tasks in robotics: \texttt{pick-place} (Fig.~\ref{fig:envs-pick_place}), \texttt{bin-picking} (Fig.~\ref{fig:envs-bin_picking}), and \texttt{door-open} (Fig.~\ref{fig:envs-door}).
In \texttt{pick-place}, the agent must move the block to the goal among a distractor object placed in the scene. In \texttt{bin-picking}, the agent must move the block from the right-side bin to the left-side bin. In \texttt{door-open}, the agent must pull on the door handle. We use scripted expert policies from the \emph{Meta World} benchmark, which compute expert actions given the object poses, to collect demonstrations in each simulated task. The agent is given $84 \times 84$ RBG image observations, the robot's end-effector position from the last two time-steps, and the distance between the robot's fingers from the last two time-steps. The actions are the desired change in the 3D-position of the end-effector and whether to open or close the gripper.

\subsection{Algorithms and Evaluation Protocol}
\label{subsec:algorithms}

The real robot manipulation policy uses the RT-1 architecture~\citep{brohan2022rt}, which tokenizes the images, text, and actions, attends over these tokens with a Transformer~\citep{vaswani2017attention}, and trains with a language-conditioned imitation learning objective. In simulation, we equip vanilla behavior cloning with several different methods for improving generalization. Specifically, we evaluate techniques for image data augmentation (random crops and random photometric distortions) and evaluate pretrained representations (CLIP~\citep{radford2021learning} and R3M~\citep{nair2022r3m}) for encoding image observations. More details on the implementation and training procedure can be found in App.~\ref{app:implementation}.

\textbf{Evaluation protocol.} On the real robot task, we evaluate the policies on two new lighting conditions, three sets of new distractor objects, three new table textures, three new backgrounds, and two new camera poses. For each factor of interest, we conduct two evaluation trials in each of the six tasks, and randomly shuffle the object and distractor positions between trials. We report the success rate averaged across the $12$ trials. To evaluate the generalization behavior of the trained policies in \emph{Factor World}, we shift the train environments by randomly sampling $100$ new values for the factor of interest, creating $100$ test environments. We report the average \textbf{generalization gap}, which is defined as $P_T - P_F$, where $P_T$ is the success rate on the train environments and $P_F$ is the new success rate under shifts to factor F. See App.~\ref{app:exp_details} for more details on our evaluation metrics.

\section{Experimental Results}
In our experiments, we aim to answer the following questions:
\begin{itemize}[leftmargin=0.5cm,noitemsep]
    \item How much does each environment factor contribute to the generalization gap? (Sec.~\ref{subsec:factors})

    \item What effects do data augmentation and pretrained representations have on the generalization performance? (Sec.~\ref{subsec:rep_aug})

    \item How do different data collection strategies, such as prioritizing visual diversity in the data, impact downstream generalization? (Sec.~\ref{subsec:collect})
\end{itemize}

\subsection{Impact of Environment Factors on Generalization}
\label{subsec:factors}

\begin{figure*}
    \centering
    \includegraphics[width=\linewidth]{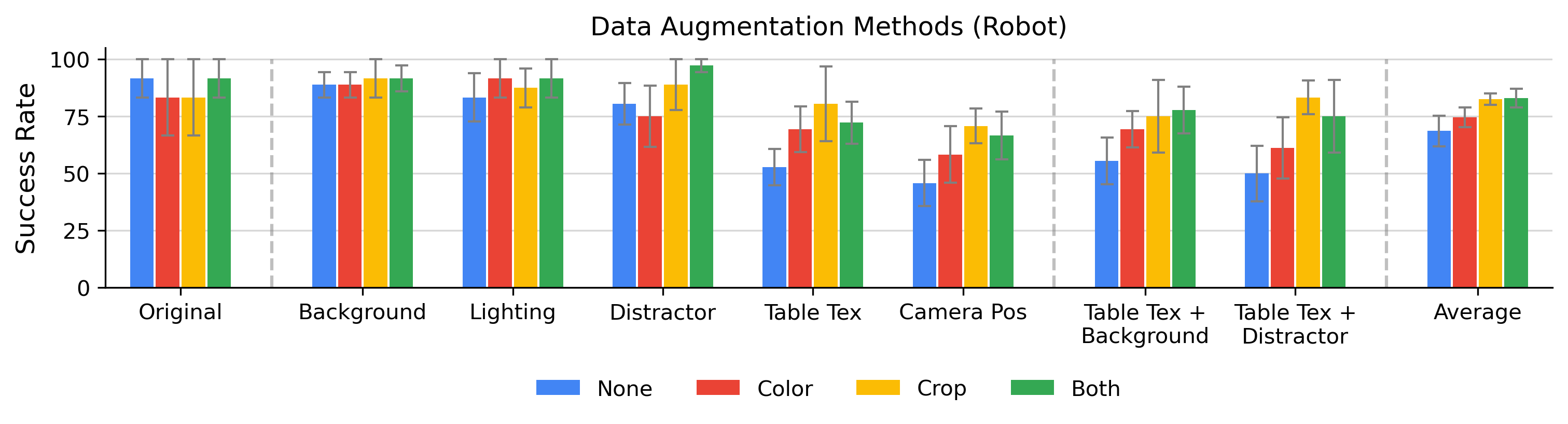}
    \vspace{-0.4cm}
    \caption{\small Performance of real-robot policies trained without data augmentation (blue), with random photometric distortions (red), with random crops (yellow), and with both (green). The results discussed in Sec.~\ref{subsec:factors} are with ``Both''. ``Original'' is the success rate on train environments, ``Background'' is the success rate when we perturb the background, ``Distractors'' is where we replace the distractors with new ones, etc. Error bars represent standard error of the mean. We also provide the average over all $7$ (sets of) factors on the far right.
    }
    \vspace{-0.2cm}
    \label{fig:robot_aug}
\end{figure*}

\textbf{Individual factors.} We begin our real robot evaluation by benchmarking the model's performance on the set of six training tasks, with and without shifts. Without shifts, the policy achieves an average success rate of $91.7\%$. Our results with shifts are presented in Fig.~\ref{fig:robot_aug}, as the set of green bars. We find that the new backgrounds have little impact on the performance ($88.9\%$), while new distractor objects and new lighting conditions have a slight effect, decreasing success rate to $80.6\%$ and $83.3\%$ respectively. Finally, changing the table texture and camera orientation causes the biggest drop, to $52.8\%$ and $45.8\%$, as the entire dataset uses a fixed head pose. Since we use the same patterned paper to introduce variations in backgrounds and table textures, we can make a direct comparison between these two factors, and conclude that new textures are harder to generalize to than new backgrounds. 

\begin{figure*}
    \centering
    \subfloat[\label{fig:sim_single_line}]{
        \includegraphics[height=1.3in,trim={7pt 0 0 0},clip]{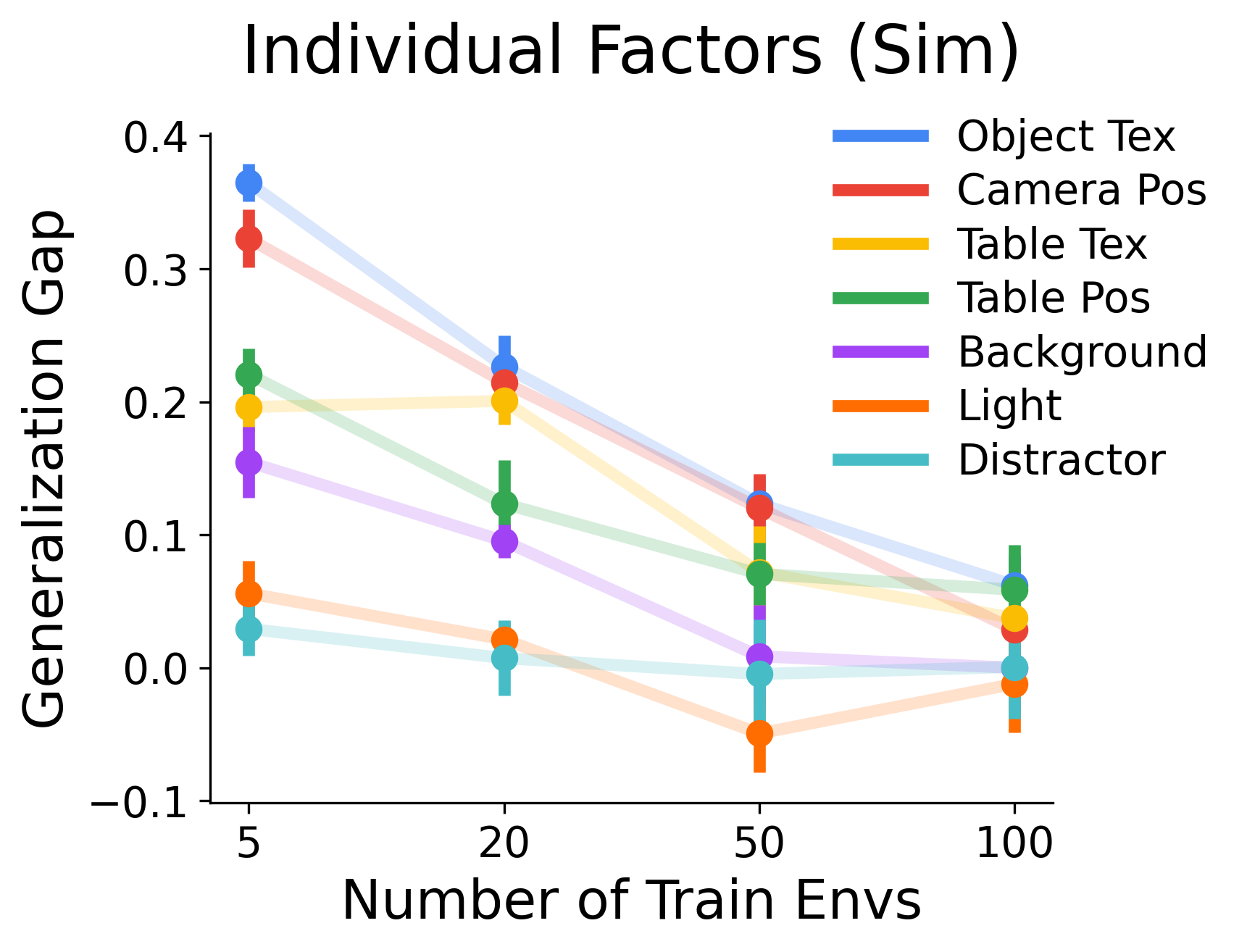}
    }
    \subfloat[\label{fig:sim_continuous}]{
        \includegraphics[height=1.3in,trim={7pt 0 0 0},clip]{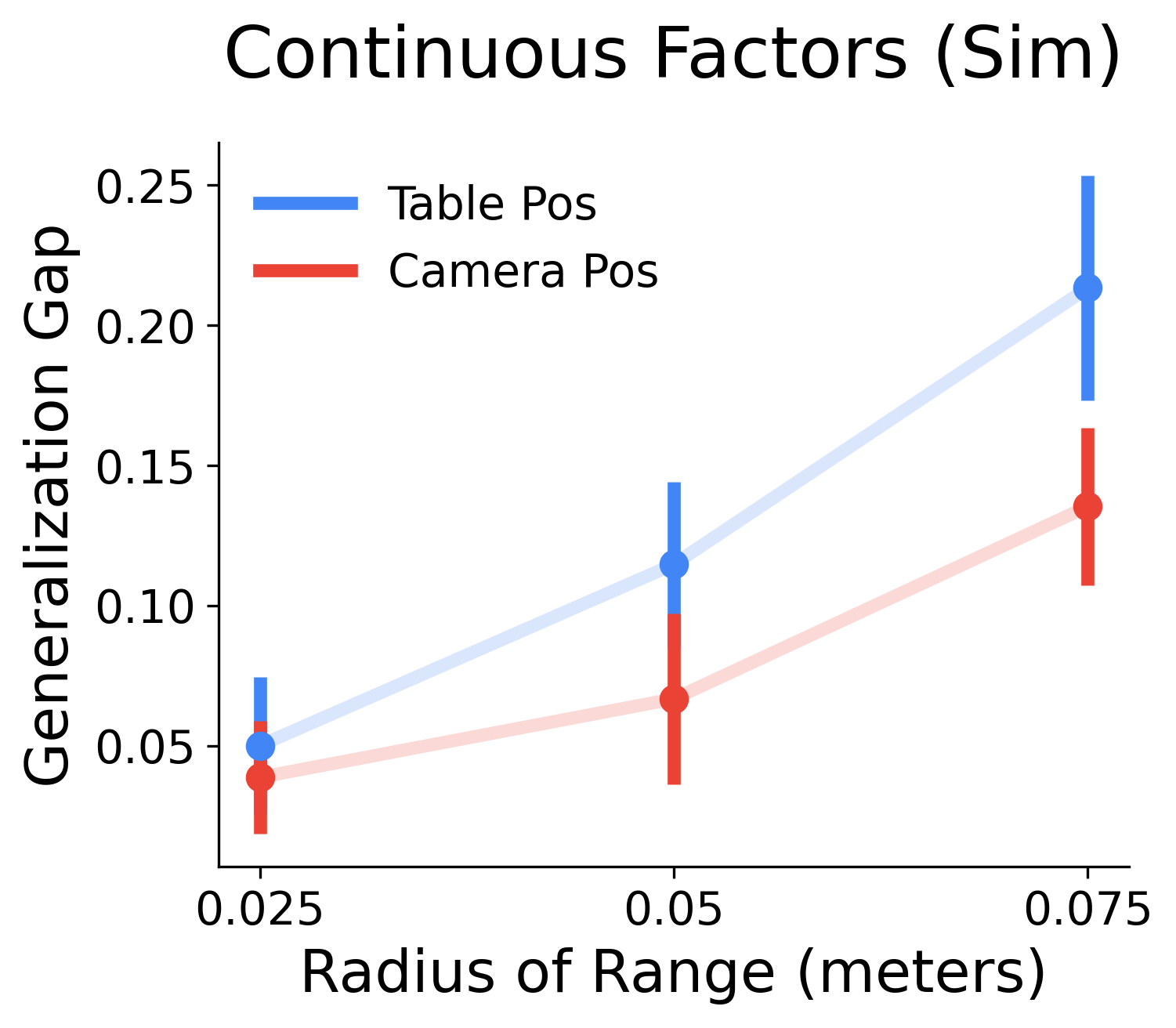}
    }
    \subfloat[\label{fig:sim_pair}]{
        \includegraphics[height=1.3in]{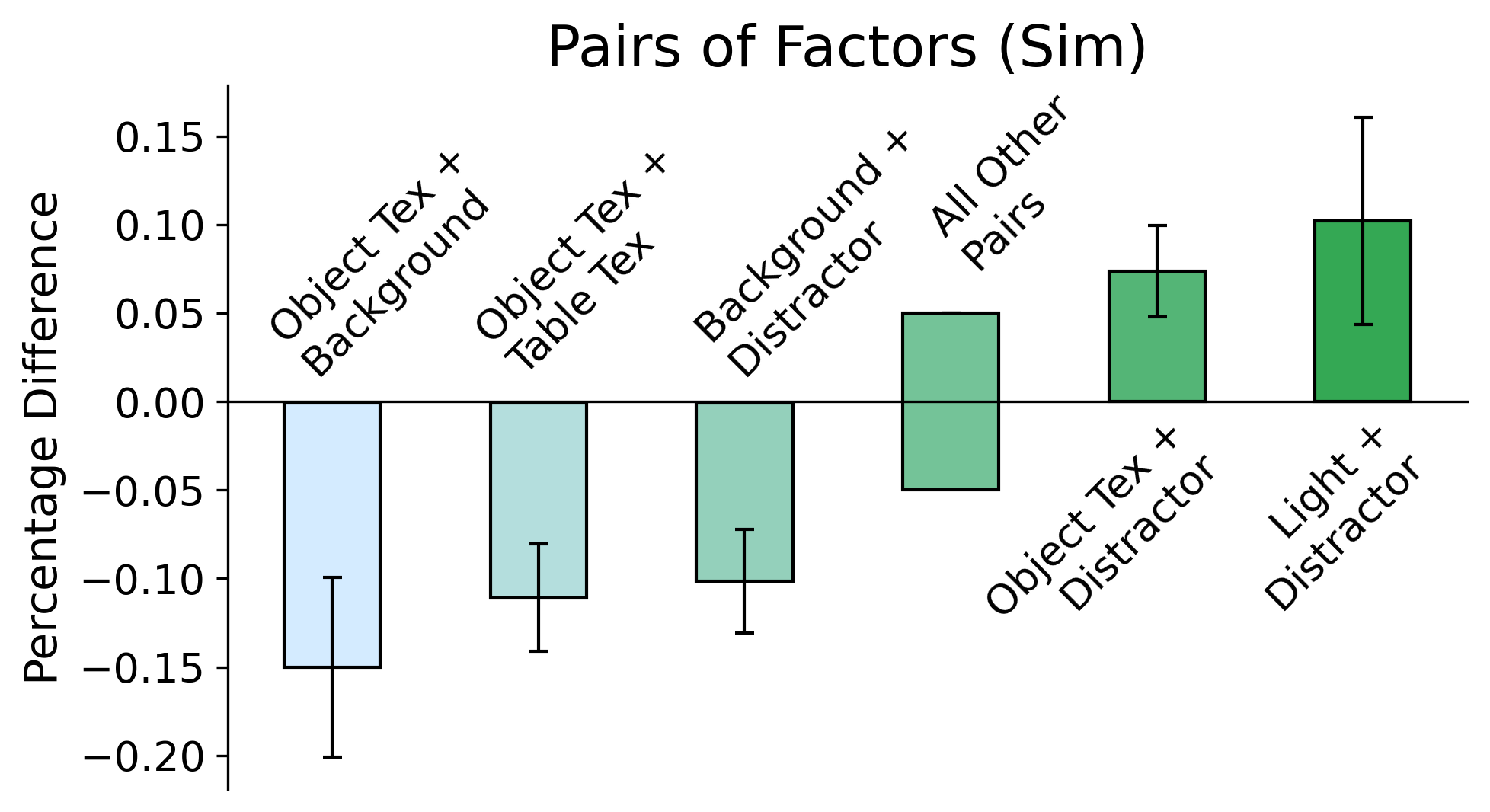}
    }
    \caption{\small (a) Generalization gap when shifts are introduced to individual factors in \emph{Factor World}. (b) Generalization gap versus the radius of the range that camera and table positions are sampled from, in \emph{Factor World}. (c) Performance on pairs of factors, reported as the percentage difference relative to the harder factor of the pair, in \emph{Factor World}. All results are averaged across the $3$ simulated tasks with $5$ seeds for each task. Error bars represent standard error of the mean.
    }
    \vspace{-0.2cm}
    \label{fig:sim_main}
\end{figure*}

Fig.~\ref{fig:sim_single_line} compares the generalization gap due to each individual factor on \emph{Factor World}. We plot this as a function of the number of training environments represented in the dataset, where an environment is parameterized by the sampled value for each factor of variation. For the continuous-valued factors, camera position and table position, we sample from the ``Narrow'' ranges (see App.~\ref{app:exp_details} for the exact range values). \textbf{Consistent across simulated and real-world results, new backgrounds, distractors, and lighting are easier factors to generalize to, while new table textures and camera positions are harder.} In \emph{Factor World}, new backgrounds are harder than distractors and lighting, in contrast to the real robot results, where they were the easiest. This may be explained by the fact that the real robot dataset contains a significant amount of background diversity, relative to the lighting and distractor factors, as described in Sec.~\ref{subsec:tasks}. In \emph{Factor World}, we additionally study object textures and table positions, including the height of the table. New object textures are about as hard to overcome as new camera positions, and new table positions are as hard as new table textures. Fortunately, the generalization gap closes significantly for \emph{all} factors, from a maximum gap of $0.4$ to less than $0.1$, when increasing the number of training environments from $5$ to $100$. 

\textbf{Pairs of factors.} Next, we evaluate performance with respect to pairs of factors to understand how they interact, i.e., whether generalization to new pairs is harder (or easier) than generalizing to one of them. On the real robot, we study the factors with the most diversity in the training dataset: table texture + distractors and table texture + background. Introducing new background textures or new distractors on top of a new table texture does not make it any harder than the new table texture alone (see green bars in Fig.~\ref{fig:robot_aug}). The success rate with new table texture + new background is $55.6\%$ and with new table texture + new distractors is $50.0\%$, comparable to the evaluation with only new table textures, which is $52.8\%$. 

In \emph{Factor World}, we evaluate all $21$ pairs of the seven factors, and report with a different metric: the success rate gap, normalized by the harder of the two factors. Concretely, this metric is defined as $\left( P_{A+B} - \min(P_A, P_B) \right) / \min(P_A, P_B)$, where $P_A$ is the success rate under shifts to factor A, $P_B$ is the success rate under shifts to factor B, and $P_{A+B}$ is the success rate under shifts to both. \textbf{Most pairs of factors do not have a compounding effect on generalization performance.} For $16$ out of the $21$ pairs, the relative percentage difference in the success rate lies between $-6\%$ and $6\%$. In other words, generalizing to the combination of two factors is not significantly harder or easier than the individual factors. In Fig.~\ref{fig:sim_pair}, we visualize the performance difference for the remaining $5$ factor pairs that lie outside of this $\left(-6\%, 6\% \right)$ range (see App.~\ref{app:exp_results} for the results for all factor pairs). Interestingly, the following factors combine synergistically, making it easier to generalize to compared to the (harder of the) individual factors: object texture + distractor and light + distractor. This result suggests that we can study these factors independently of one another, and improvements with respect to one factor may carry over to scenarios with multiple factor shifts.

\textbf{Continuous factors.} The camera position and table position factors are continuous, unlike the other factors which are discrete, hence the generalization gap with respect to these factors will depend on the range that we train and evaluate on. We aim to understand how much more difficult training and generalizing to a wider range of values is, by studying the gap with the following range radii: $0.025$, $0.050$, ad $0.075$ meters. For both camera-position and table-position factors, as we linearly increase the radius, the generalization gap roughly doubles (see Fig.~\ref{fig:sim_continuous}). This pattern suggests: (1) performance can be dramatically improved by keeping the camera and table position as constant as possible, and (2) generalizing to wider ranges may require significantly more diversity, i.e., examples of camera and table positions in the training dataset. However, in Sec.~\ref{subsec:rep_aug}, we see that existing methods can address the latter issue to some degree.

\subsection{Effect of Data Augmentation and Pretrained Representations}
\label{subsec:rep_aug}

\begin{wrapfigure}{r}{0.3\linewidth}
    \centering
    \vspace{-0.5cm}
    \includegraphics[width=\linewidth]{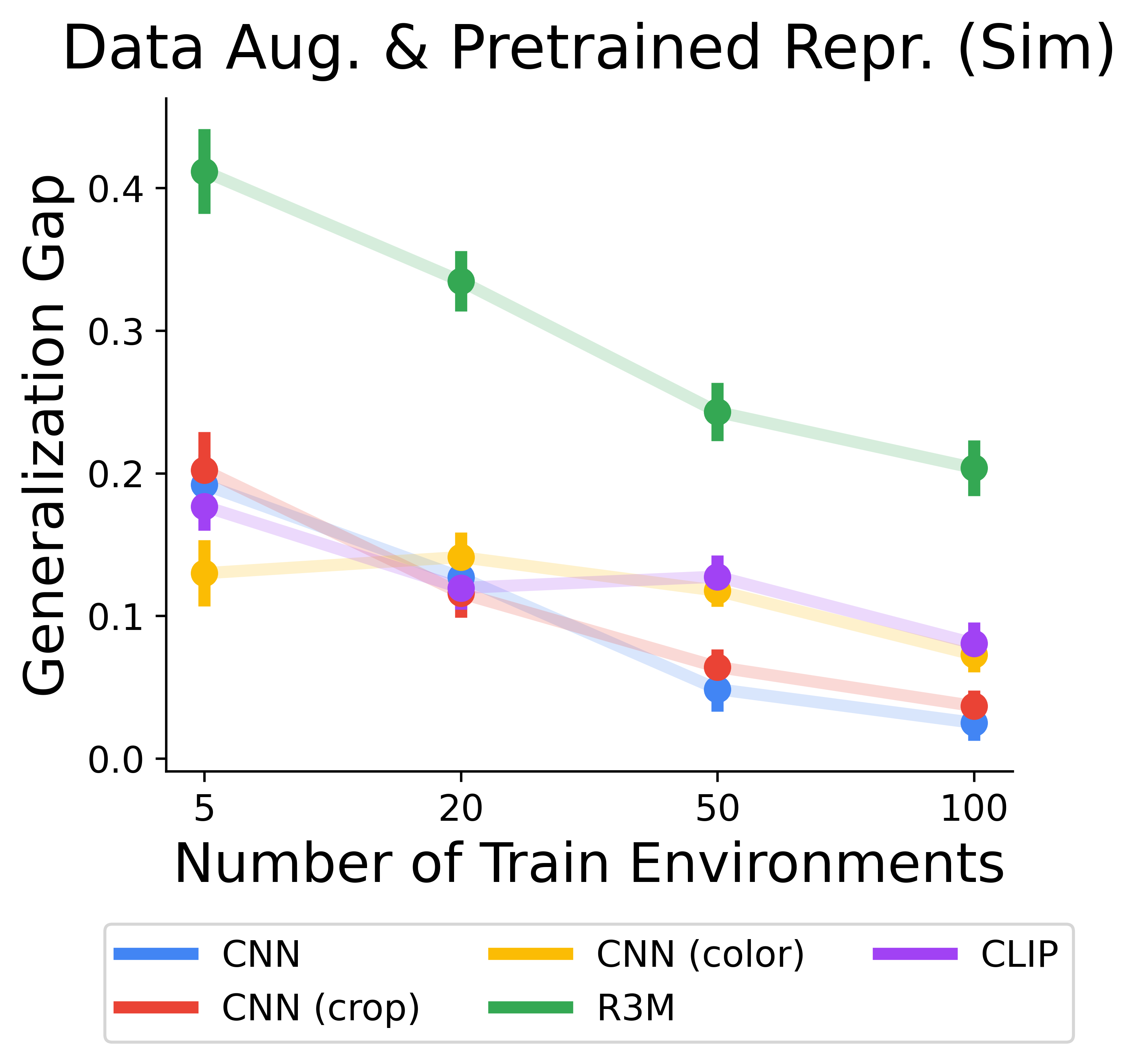}
    \caption{\small Generalization gap with data augmentations and pretrained representations in \emph{Factor World}. Lower is better. Results are averaged across the $7$ factors, $3$ tasks, and $5$ seeds for each task.
    }
    \label{fig:sim_method_avg}
    \vspace{-0.35cm}
\end{wrapfigure}
\textbf{The impact of data augmentation under individual factor shifts.} We study two forms of augmentation: (1) random crops and (2) random photometric distortions. The photometric distortion randomly adjusts the brightness, saturation, hue, and contrast of the image, and applies random cutout and random Gaussian noise. Fig.~\ref{fig:robot_aug} and Fig.~\ref{fig:sim_method_avg} show the results for the real robot and \emph{Factor World} respectively. On the robot, \textbf{crop augmentation improves generalization along multiple environment factors, most significantly to new camera positions and new table textures}. While the improvement on a spatial factor like camera position is intuitive, we find the improvement on a non-spatial factor like table texture surprising. More in line with our expectations, the photometric distortion augmentation improves the performance on texture-based factors like table texture in the real robot environment and object, table and background in the simulated environment (see App.~\ref{app:exp_results} for \emph{Factor World} results by factor).

\textbf{The impact of pretrained representations under individual factor shifts.} We study two pretrained representations: (1) R3M~\citep{nair2022r3m} and (2) CLIP~\citep{radford2021learning}. While these representations are trained on non-robotics datasets, policies trained on top of them have been shown to perform well in robotics environments from a small amount of data. However, while they achieve good performance on training environments (see Fig.~\ref{fig:sim_success_rate} in App.~\ref{app:exp_results}), they struggle to generalize to new but similar environments, leaving a large generalization gap across many factors (see Fig.~\ref{fig:sim_method_avg}). Though, CLIP does improve upon a trained-from-scratch CNN with new object textures (Fig.~\ref{fig:sim_method}; first row, fourth plot).

\subsection{Investigating Different Strategies for Data Collection}
\label{subsec:collect}

\begin{figure*}
    \centering
    \includegraphics[width=\linewidth]{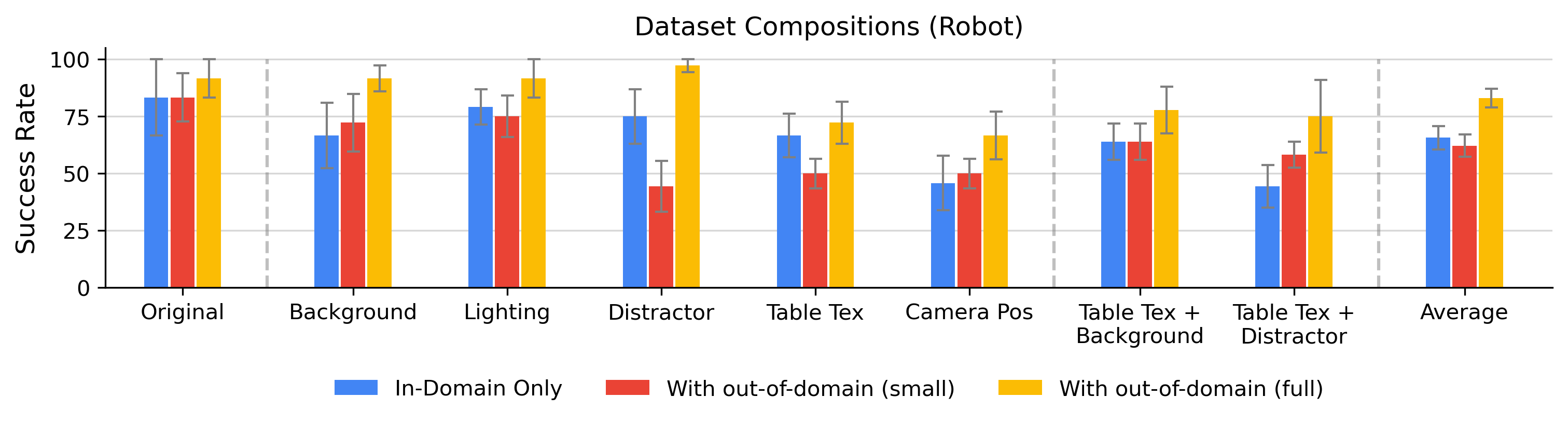}
    \caption{\small Performance of real-robot policies trained with in-domain data only (blue), a small version of the in- and out-of-domain dataset (red), and the full version of the in- and out-of-domain dataset (yellow). Error bars represent standard error of the mean. We also provide the average over all $7$ (sets of) factors on the far right.}
    \label{fig:robot_dataset}
\end{figure*}

\textbf{Augmenting visual diversity with out-of-domain data.} As described in Sec.~\ref{subsec:tasks}, our real robot dataset includes demonstrations collected from other domains and tasks like opening a fridge and operating a cereal dispenser. Only $35.2\%$ of the $115$K demonstrations are collected in the same domain as our evaluations. While the remaining demonstrations are out of domain and focus on other skills such as drawer manipulation, they add visual diversity, such as new objects and new backgrounds, and demonstrate robotic manipulation behavior, unlike the data that R3M and CLIP pretrain on. We consider the dataset with only in-domain data, which we refer to as \texttt{In-domain only}. In Fig.~\ref{fig:robot_dataset}, we compare \texttt{In-domain only} (blue) to the full dataset, which we refer to as \texttt{With out-of-domain (full)} (yellow). While the performance on the original six training tasks is comparable, the success rate of the \texttt{In-domain only} policy drops significantly across the different environment shifts, and the \texttt{With out-of-domain (full)} policy is more successful across the board. \textbf{Unlike representations pretrained on non-robotics datasets (Sec.~\ref{subsec:rep_aug}), out-of-domain robotics data can improve in-domain generalization.}

\textbf{Prioritizing visual diversity with out-of-domain data.} Finally, we consider a uniformly subsampled version of the \texttt{With out-of-domain (full)} dataset, which we refer to as \texttt{With out-of-domain (small)}. \texttt{With out-of-domain (small)} has the same number of demonstrations as \texttt{In-domain only}, allowing us to directly compare whether the in-domain data or out-of-domain data is more valuable. We emphasize that \texttt{With out-of-domain (small)} has significantly fewer in-domain demonstrations of the ``pick'' skill than \texttt{In-domain only}. Intuitively, one would expect the in-domain data to be more useful. However, in Fig.~\ref{fig:robot_dataset}, we see that the \texttt{With out-of-domain (small)} policy (red) performs comparably with the \texttt{In-domain only} policy (blue) across most of the factors. The main exception is scenarios with new distractors, where the \texttt{In-domain only} policy has a $75.0\%$ success rate while the \texttt{With out-of-domain (small)} policy is successful in $44.4\%$ of the trials.

\section{Discussion}

\textbf{Summary.} In this work, we studied the impact of different environmental variations on generalization performance. We determined an ordering of the environment factors in terms of generalization difficulty, that is consistent across simulation and our real robot setup, and quantified the impact of different solutions like data augmentation. Notably, many of the solutions studied were developed for computer vision tasks like image classification. While some of them transferred well to the robotic imitation learning setting, it may be fruitful to develop algorithms that prioritize this setting and its unique considerations, including the sequential nature of predictions and the often continuous, multi-dimensional action space in robotic setups. We hope this work encourages researchers to develop solutions that target the specific challenges in robotic generalization identified by our work.

\textbf{Limitations.} There are limitations to our study, which focuses on a few, but representative, robotic tasks and environment factors in the imitation setting. Our real-robot experiments required conducting a total number of 1440 evaluations over all factor values, tasks, and methods, and it is challenging to increase the scope of the study because of the number of experiments required. Fortunately, future work can utilize our simulated benchmark \emph{Factor World} to study additional tasks, additional factors, and generalization in the reinforcement learning setting. We also saw that the performance on training environments slightly degrades as we trained on more varied environments (see Fig.~\ref{fig:sim_success_rate} in App.~\ref{app:exp_results}). Based on this observation, studying higher-capacity models, such as those equipped with ResNet or Vision Transformer architectures, which can likely fit these varied environments better, would be a fruitful next step.


\clearpage
\acknowledgments{We thank Yao Lu, Kaylee Burns, and Evan Liu for helpful discussions and feedback, and Brianna Zitkovich and Jaspiar Singh for their assistance in the robot evaluations. This work was supported in part by ONR grants N00014-21-1-2685 and N00014-22-1-2621.}


\bibliography{references}  

\newpage
\appendix
\section{Appendix}

\subsection{Experimental Details}
\label{app:exp_details}

In this section, we provide additional details on the experimental setup and evaluation metrics.

\subsubsection{Experimental Setup}

\begin{figure*}[ht]
    \centering
    \captionsetup[subfloat]{labelfont=scriptsize,textfont=scriptsize}
    \subfloat[Pepsi, water bottle\label{fig:pick_pepsi}]{
        \includegraphics[width=0.185\textwidth]{original.jpg}
    }
    \subfloat[Blue chip bag\label{fig:pick_blue_chip}]{
      \includegraphics[width=0.185\textwidth]{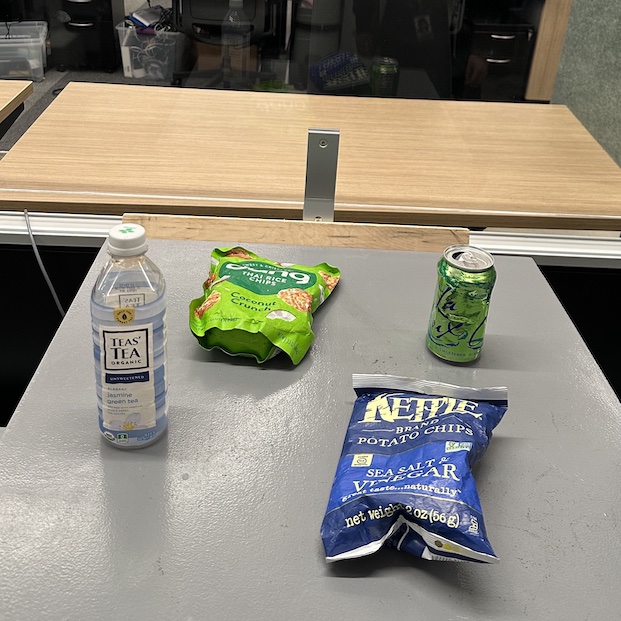}
    }
    \subfloat[Blue plastic bottle\label{fig:pick_blue_bottle}]{
      \includegraphics[width=0.185\textwidth]{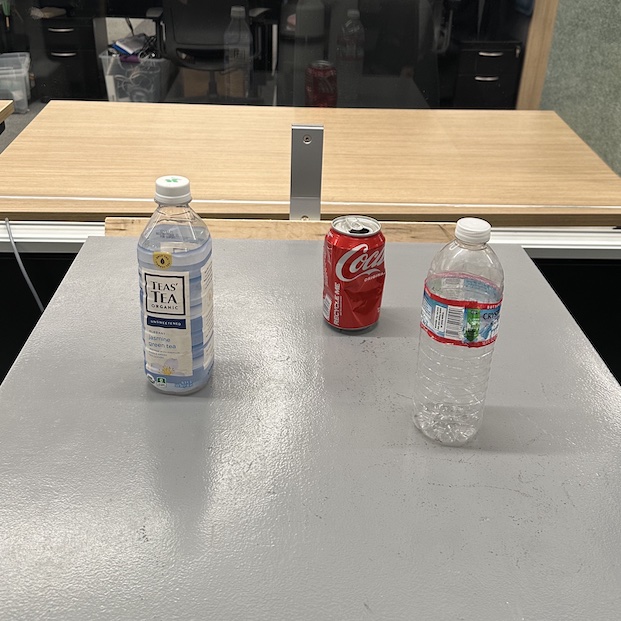}
    }
    \subfloat[Green jalapeno chip bag\label{fig:pick_green_jalapeno}]{
      \includegraphics[width=0.185\textwidth]{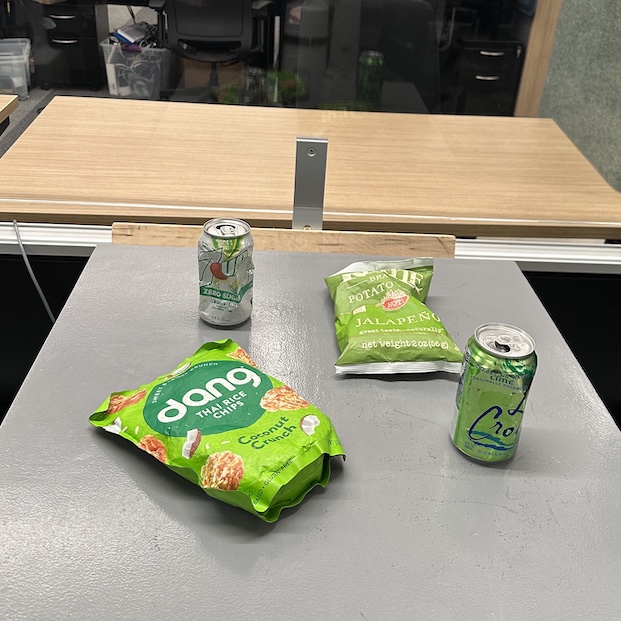}
    }
    \subfloat[Oreo\label{fig:pick_oreo}]{
      \includegraphics[width=0.185\textwidth]{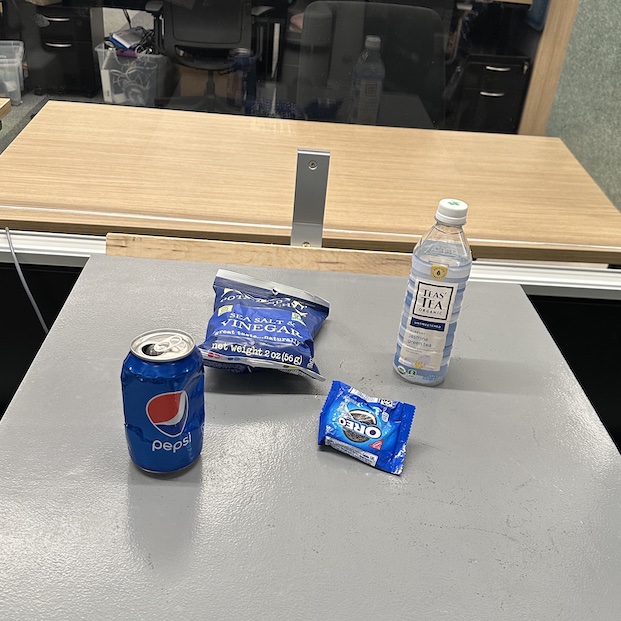}
    }
    \caption{\small The six pick tasks in our real robot evaluations.}
    \label{fig:pick_tasks}
\end{figure*}

\emph{Real robot tasks.} We define six real-world picking tasks: pepsi can, water bottle, blue chip bag, green jalapeno chip bag, and oreo, which are visualized in Fig.~\ref{fig:pick_tasks}.

\begin{figure*}[h]
\centering
\subfloat[Object position\label{fig:factors-object_pos}]{
  \includegraphics[width=0.18\textwidth,trim={20pt 0 50pt 0},clip]{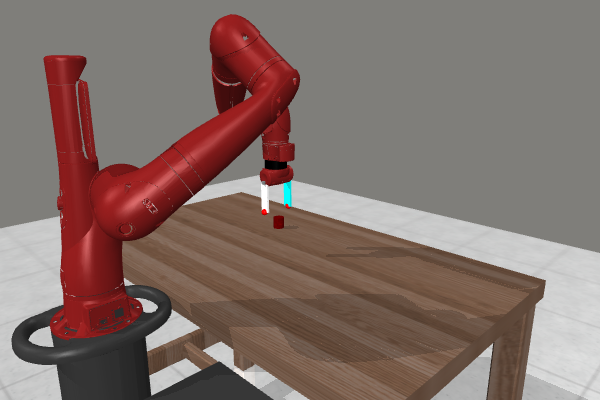}\hspace{1pt}
  \includegraphics[width=0.18\textwidth,trim={20pt 0 50pt 0},clip]{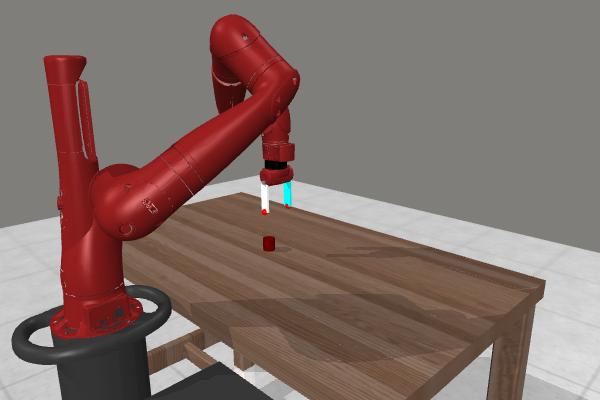}
}
\hspace{0.2cm}
\subfloat[Initial arm position\label{fig:factors-arm_pos}]{
  \includegraphics[width=0.18\textwidth,trim={20pt 0 50pt 0},clip]{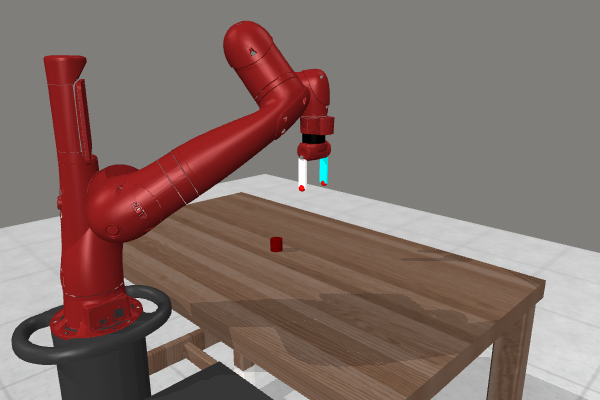}\hspace{1pt}
  \includegraphics[width=0.18\textwidth,trim={20pt 0 50pt 0},clip]{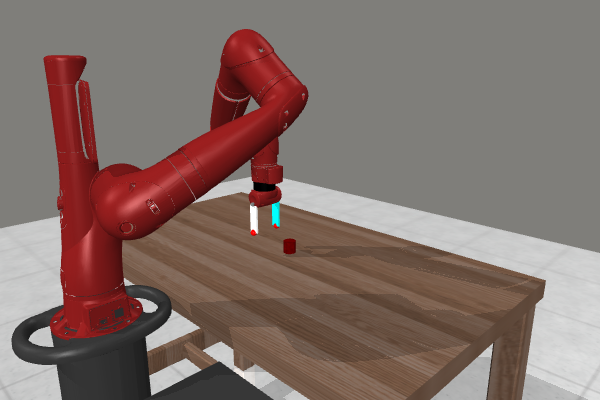}
}

\subfloat[Camera position\label{fig:factors-camera_pos}]{
  \includegraphics[width=0.18\textwidth,trim={20pt 0 50pt 0},clip]{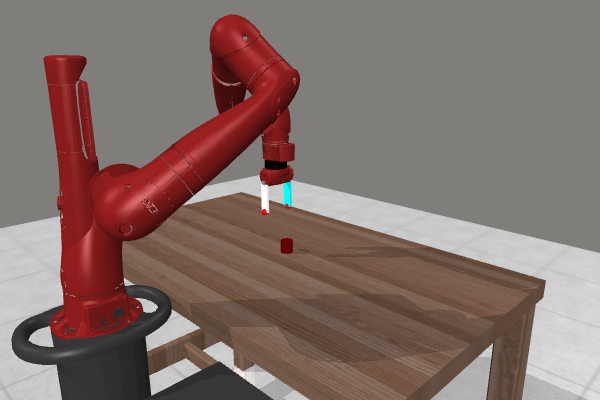}\hspace{1pt}
  \includegraphics[width=0.18\textwidth,trim={20pt 0 50pt 0},clip]{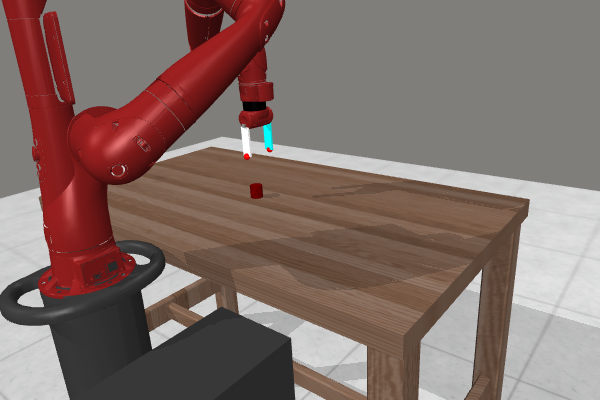}
}
\hspace{0.2cm}
\subfloat[Table position\label{fig:factors-table_pos}]{
  \includegraphics[width=0.18\textwidth,trim={20pt 0 50pt 0},clip]{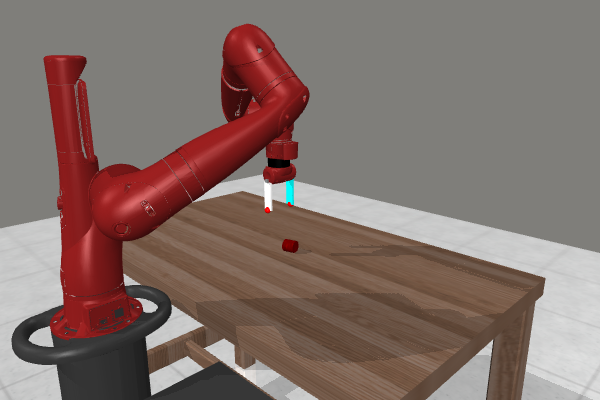}\hspace{1pt}
  \includegraphics[width=0.18\textwidth,trim={20pt 0 50pt 0},clip]{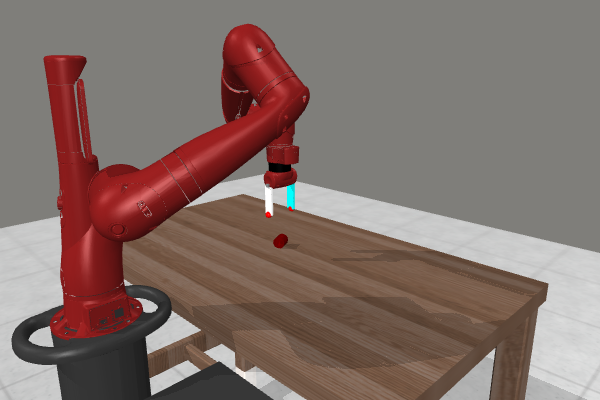}
}

\subfloat[Object size\label{fig:factors-object_size}]{
  \includegraphics[width=0.18\textwidth,trim={20pt 0 50pt 0},clip]{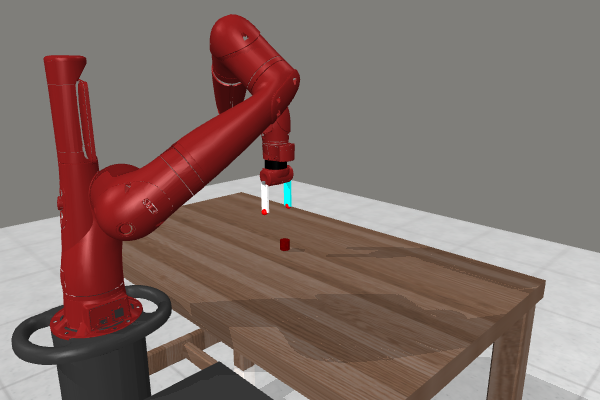}\hspace{1pt}
  \includegraphics[width=0.18\textwidth,trim={20pt 0 50pt 0},clip]{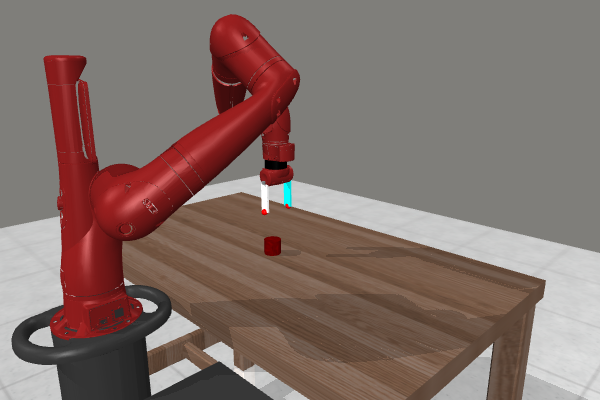}
}
\hspace{0.2cm}
\subfloat[Object texture\label{fig:factors-object_texture}]{
  \includegraphics[width=0.18\textwidth,trim={20pt 0 50pt 0},clip]{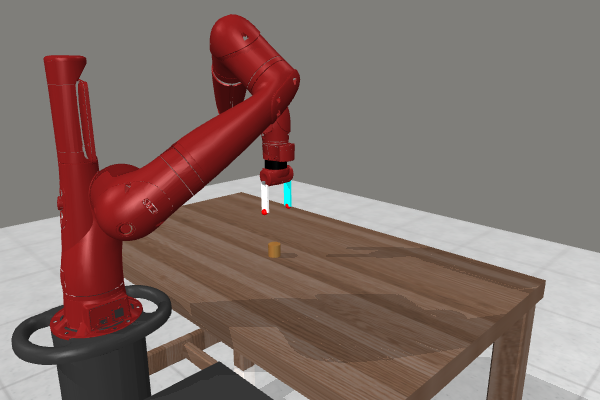}\hspace{1pt}
  \includegraphics[width=0.18\textwidth,trim={20pt 0 50pt 0},clip]{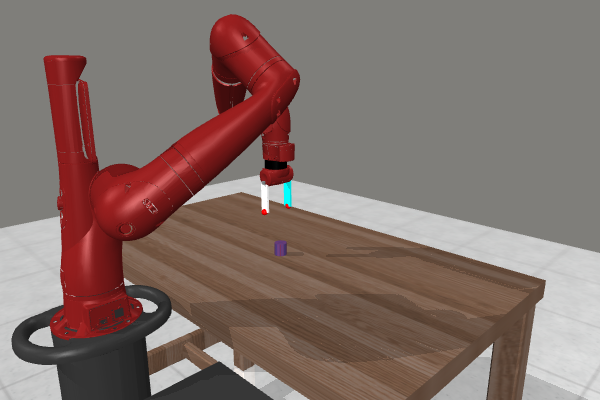}
}

\subfloat[Distractor objects \& positions\label{fig:factors-distractor}]{
  \includegraphics[width=0.18\textwidth,trim={20pt 0 50pt 0},clip]{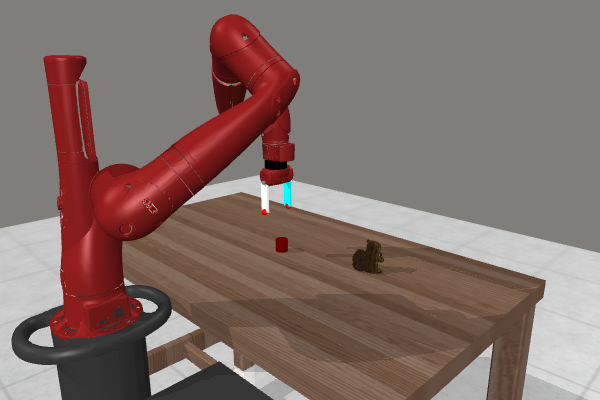}\hspace{1pt}
  \includegraphics[width=0.18\textwidth,trim={20pt 0 50pt 0},clip]{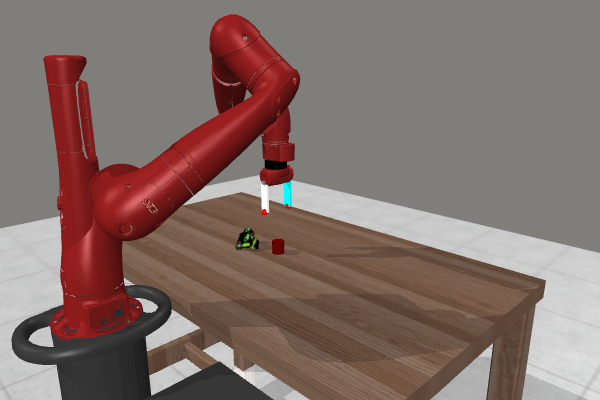}
}
\hspace{0.2cm}
\subfloat[Floor texture\label{fig:factors-floor_texture}]{
  \includegraphics[width=0.18\textwidth,trim={20pt 0 50pt 0},clip]{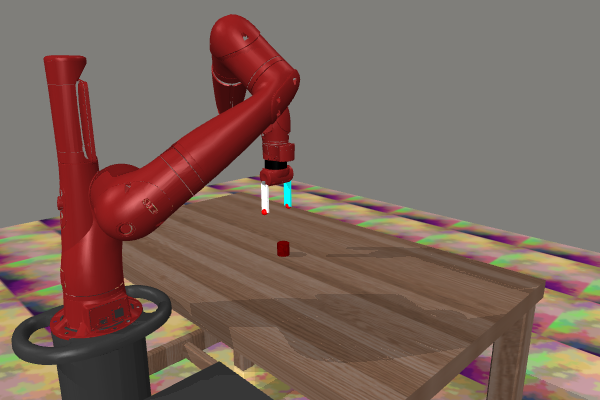}\hspace{1pt}
  \includegraphics[width=0.18\textwidth,trim={20pt 0 50pt 0},clip]{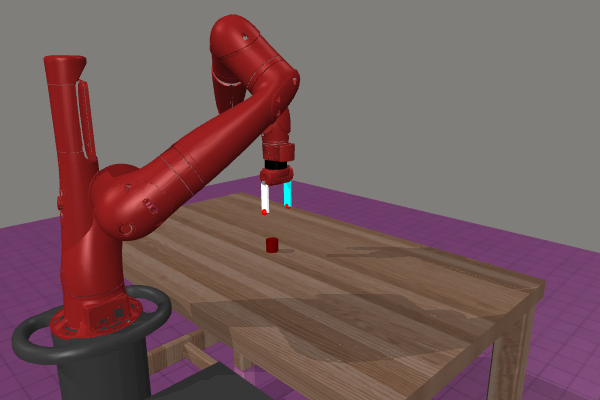}
}

\subfloat[Table texture\label{fig:factors-table_texture}]{
  \includegraphics[width=0.18\textwidth,trim={20pt 0 50pt 0},clip]{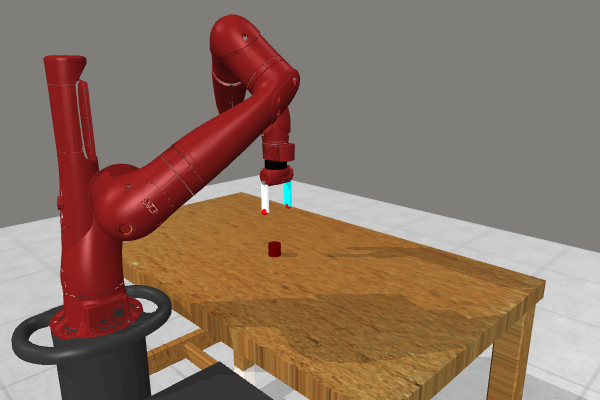}\hspace{1pt}
  \includegraphics[width=0.18\textwidth,trim={20pt 0 50pt 0},clip]{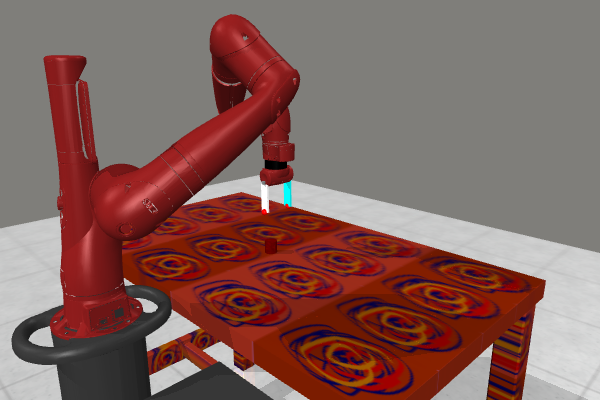}
}
\hspace{0.2cm}
\subfloat[Lighting\label{fig:factors-lighting}]{
  \includegraphics[width=0.18\textwidth,trim={20pt 0 50pt 0},clip]{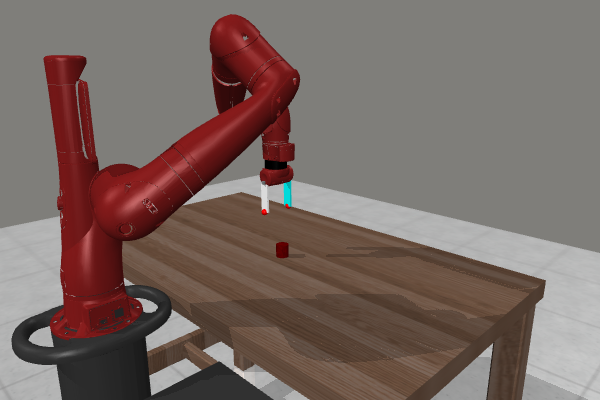}\hspace{1pt}
  \includegraphics[width=0.18\textwidth,trim={20pt 0 50pt 0},clip]{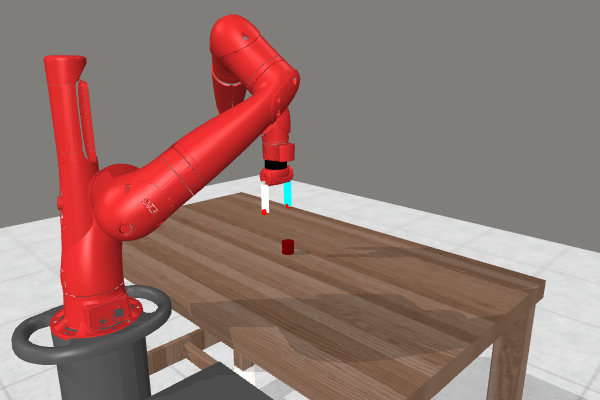}
}
\caption{\small The $11$ factors of variation implemented into \emph{Factor World}, depicted for the \texttt{pick-place} environment. Videos are available at: \url{https://sites.google.com/view/factor-envs}}
\label{fig:factors-of-variation}
\end{figure*}

\emph{Factor World.} The factors of variation implemented into \emph{Factor World} are enumerated in Fig.~\ref{fig:factors-of-variation}. In Table~\ref{tbl:continuous}, we specify the ranges of the continuous-valued factors.

\setlength{\tabcolsep}{.11cm}
\begin{table}[h]
\centering
\small
\begin{tabular}{lllll}
    \toprule
    Factor & Parameters & Narrow & Medium & Wide \\
    \midrule
    \multirow{2}{*}{Object position} & X-position & $[-0.05, 0.05]$ & $[-0.1, 0.1]$ & - \\
                              & Y-position & $[-0.05, 0.05]$ & $[-0.075, 0.075]$ & - \\
    \midrule
    \multirow{7}{*}{Camera position} & X-position & $[-0.025, 0.025]$ & $[-0.05, 0.05]$ & $[-0.075, 0.075]$  \\
                              & Y-position & $[-0.025, 0.025]$ & $[-0.05, 0.05]$ & $[-0.075, 0.075]$ \\
                              & Z-position & $[-0.025, 0.025]$ & $[-0.05, 0.05]$ & $[-0.075, 0.075]$  \\
                              & $q_1$ & $[-0.025, 0.025]$ & $[-0.05, 0.05]$ & $[-0.075, 0.075]$  \\
                              & $q_2$ & $[-0.025, 0.025]$ & $[-0.05, 0.05]$ & $[-0.075, 0.075]$  \\
                              & $q_3$ & $[-0.025, 0.025]$ & $[-0.05, 0.05]$ & $[-0.075, 0.075]$  \\
                              & $q_4$ & $[-0.025, 0.025]$ & $[-0.05, 0.05]$ & $[-0.075, 0.075]$  \\
    \midrule
    \multirow{3}{*}{Table position} & X-position & $[-0.025, 0.025]$ & $[-0.05, 0.05]$ & $[-0.075, 0.075]$  \\
                              & Y-position & $[-0.025, 0.025]$ & $[-0.05, 0.05]$ & $[-0.075, 0.075]$  \\
                              & Z-position & $[-0.025, 0.025]$ & $[-0.05, 0.025]$ & $[-0.05, 0.025]$ \\
    \bottomrule
\end{tabular}
\vspace{0.1cm}
\caption{\small Range for each continuous factor in meters. As a point of comparison for the position-based factors, the table in the environment measures at $0.7 m \times 0.4 m$. 
}
\label{tbl:continuous}
\end{table}

\subsubsection{Dataset Details}

\emph{Factor World datasets.} In the \texttt{pick-place} task, we collect datasets of $2000$ demonstrations, across $N = 5, 20, 50, 100$ training environments. A training environment is parameterized by a collection of factor values, one for each environment factor. We collect datasets of $1000$ demonstrations for \texttt{bin-picking} and \texttt{door-open}, which we empirically found to be easier than the \texttt{pick-place} task.

\subsubsection{Evaluation Metrics}

\emph{Generalization gap.} Besides the success rate, we also measure the generalization gap which is defined as the difference between the performance on the train environments and the performance on the test environments. The test environments have the same setup as the train environments, except $1$ (or $2$ in the factor pair experiments) of the factors is assigned a new value. For example, in Fig.~\ref{fig:real_single_factor_bar}, `Background' represents the change in success rate when introducing new backgrounds to the train environments.

\emph{Percentage difference.} When evaluating a pair of factors, we report the percentage difference with respect to the harder of the two factors. Concretely, this metric is computed as $\left( p_{A+B} - \min(p_A, p_B) \right) / \min(p_A, p_B)$, where $p_A$ is the success rate under shifts to factor A, $p_A$ is the success rate under shifts to factor B, and $p_{A+B}$ is the success rate under shifts to both. 

\subsection{Implementation and Training Details}
\label{app:implementation}

In this section, we provide additional details on the implementation and training of all models.

\subsubsection{RT-1}
\emph{Behavior cloning.} We follow the RT-1 architecture that uses tokenized image and language inputs with a categorical cross-entropy objective for tokenized action outputs. The model takes as input a natural language instruction along with the $6$ most recent RGB robot observations, and then feeds these through pre-trained language and image encoders (Universal Sentence Encoder~\citep{cer2018universal} and EfficientNet-B3~\citep{pmlr-v97-tan19a}, respectively). These two input modalities are fused with FiLM conditioning, and then passed to a TokenLearner~\citep{ryoo2021tokenlearner} spatial attention module to reduce the number of tokens needed for fast on-robot inference. Then, the network contains 8 decoder only self-attention Transformer layers, followed by a dense action decoding MLP layer. Full details of the RT-1 architecture that we follow can be found in \citep{brohan2022rt}.

\emph{Data augmentations.} Following the image augmentations introduced in Qt-Opt~\citep{kalashnikov2018qtopt}, we perform two main types of visual data augmentation during training only: visual disparity augmentations and random cropping. For visual disparity augmentations, we adjust the brightness, contrast, and saturation by sampling uniformly from [-0.125, 0.125], [0.5, 1.5], and [0.5, 1.5] respectively. For random cropping, we subsample the full-resolution camera image to obtain a $300 \times 300$ random crop. Since RT-1 uses a history length of 6, each timestep is randomly cropped independently. 

\emph{Pretrained representations.} Following the implementation in RT-1, we utilize an EfficientNet-B3 model pretrained on ImageNet~\citep{pmlr-v97-tan19a} for image tokenization, and the Universal Sentence Encoder~\citep{cer2018universal} language encoder for embedding natural language instructions. The rest of the RT-1 model is initialized from scratch.

\subsubsection{Factor World}

\emph{Behavior cloning.} Our behavior cloning policy is parameterized by a convolutional neural network with the same architecture as in~\citep{sekar2020planning} and in~\citep{xing2021kitchenshift}: there are four convolutional layers with $32$, $64$, $128$, and $128$ $4 \times 4$ filters, respectively. The features are then flattened and passed through a linear layer with output dimension of $128$, LayerNorm, and Tanh activation function. The policy head is parameterized as a three-layer feedforward neural network with $256$ units per layer. All policies are trained for $100$ epochs.

\emph{Data augmentations.} In our simulated experiments, we experiment with shift augmentations (analogous to the crop augmentations the real robot policy trains with) from~\citep{yarats2021mastering}: we first pad each side of the $84 \times 84$ image by 4 pixels, and then select a random $84 \times 84$ crop. We also experiment with color jitter augmentations (analogous to the photometric distortions studied for the real robot policy), which is implemented in torchvision. The brightness, contrast, saturation, and hue factors are set to $0.2$. The probability that an image in the batch is augmented is $0.3$. All policies are trained for $100$ epochs.

\emph{Pretrained representations.} We use the ResNet50 versions of the publicly available R3M and CLIP representations. We follow the embedding with a BatchNorm, and the same policy head parameterization: three feedforward layers with $256$ units per layer. All policies are trained for $100$ epochs.

\subsection{Additional Experimental Results in  \emph{Factor World}}
\label{app:exp_results}
In this section, we provide additional results from our simulated experiments in \emph{Factor World}.

\subsubsection{Simulation: Data Augmentation and Pretrained Representations}

\begin{figure*}
    \centering
    \includegraphics[width=\linewidth]{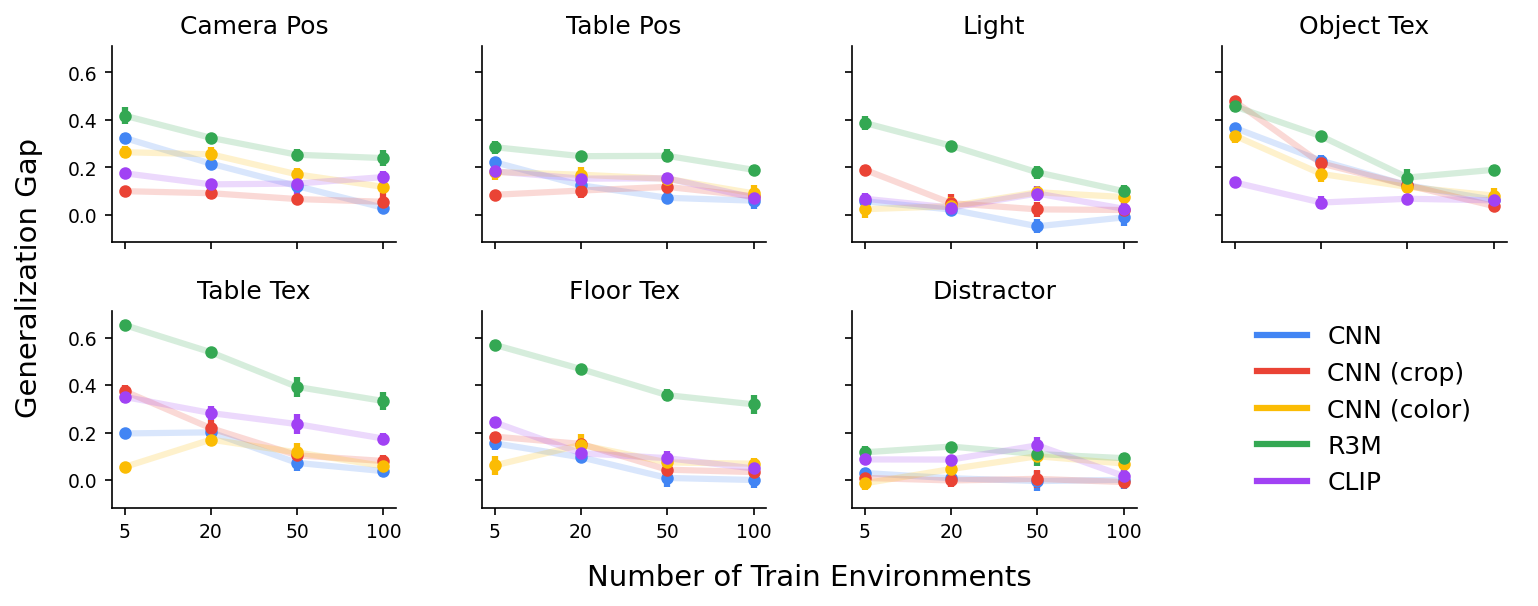}
    \vspace{-0.5cm}
    \caption{\small Generalization gap for different data augmentations and pretrained representations in \emph{Factor World}. Subplots share the same x- and y-axes. Results are averaged across the $3$ simulated tasks with $5$ seeds for each task. Error bars represent standard error of the mean.
    }
    \label{fig:sim_method}
\end{figure*}

We report the performance of the data augmentation techniques and pretrained representations by factor in Fig.~\ref{fig:sim_method}. The policy trained with the R3M representation fails to generalize well to most factors of variation, with the exception of new distractors. We also see that the policy trained with the CLIP representation performs similarly to the model trained from scratch (CNN) across most factors, except for new object textures on which CLIP outperforms the naive CNN.

\subsubsection{Simulation: Factor Pairs}
\begin{figure}
    \centering
    \includegraphics[width=0.8\linewidth]{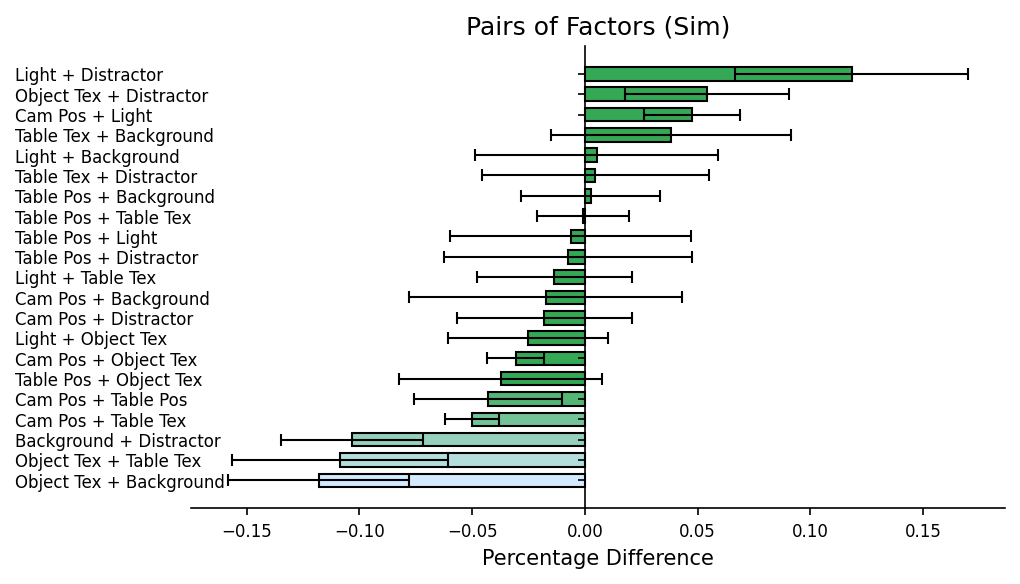}
    \caption{\small Generalization gap on all pairs of factors, reported as the percentage difference relative to the harder factor of the pair. Results are averaged across the $3$ simulated tasks with $5$ seeds for each task.}
    \label{fig:sim_pair_all}
\end{figure}

In Fig.~\ref{fig:sim_pair_all}, we report the results for all factor pairs, a partial subset of which was visualized in Fig.~\ref{fig:sim_pair}. In Fig.~\ref{fig:sim_pair}, we selected the pairs with the highest magnitude percentage difference, excluding the pairs with error bars that overlap with zero.

\subsubsection{Simulation: Success Rates}
\begin{figure*}
    \centering
    \includegraphics[width=0.9\textwidth]{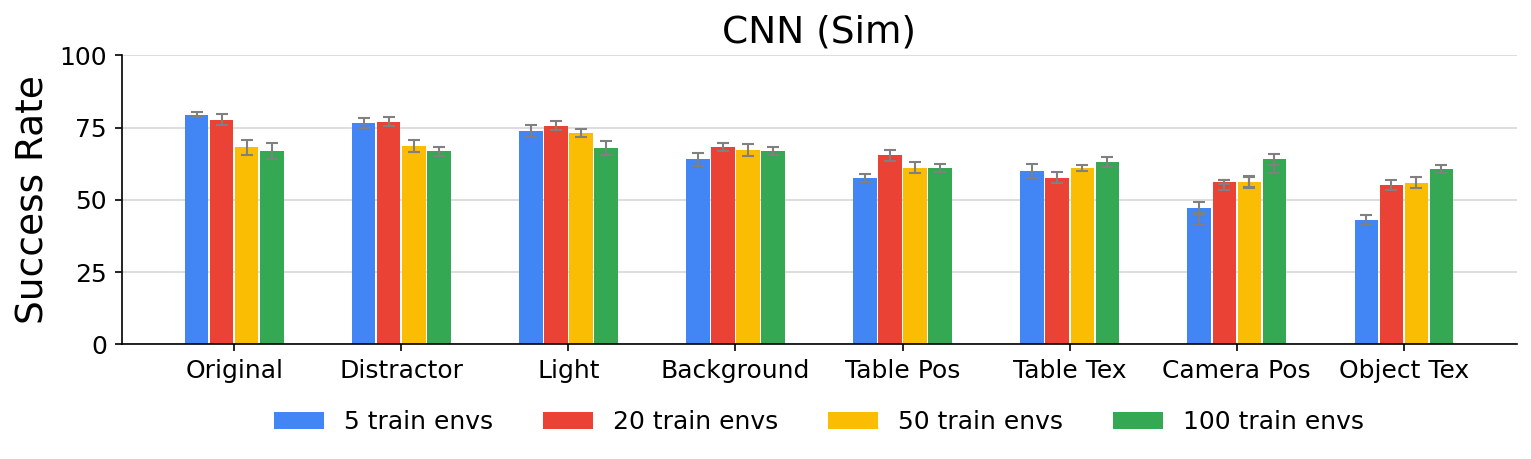}\\
    \includegraphics[width=0.9\textwidth]{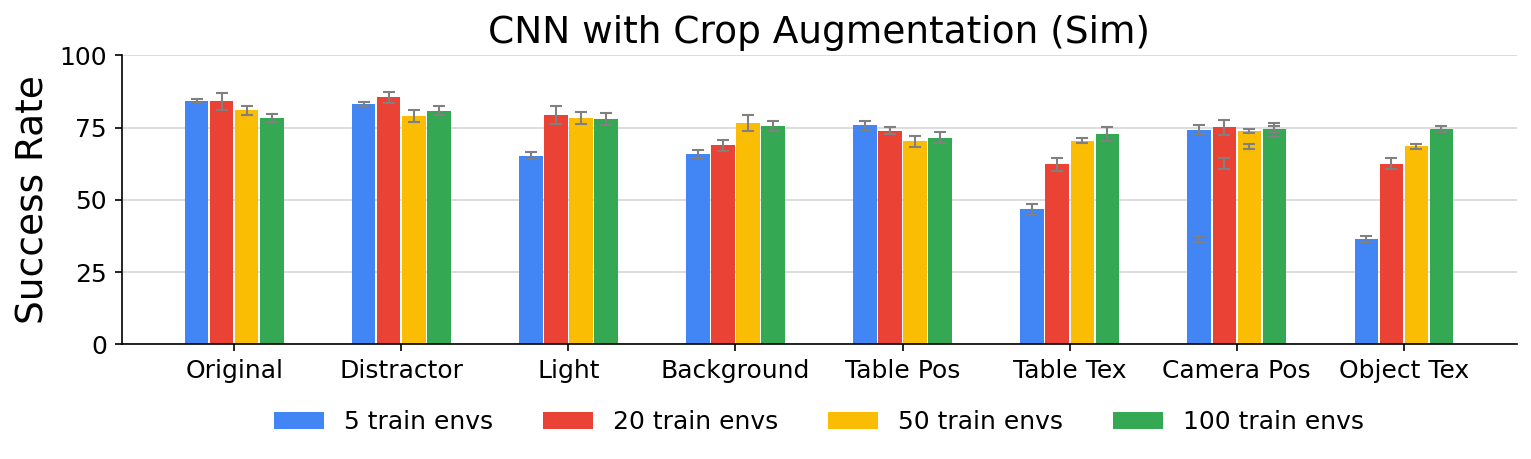}\\
    \includegraphics[width=0.9\textwidth]{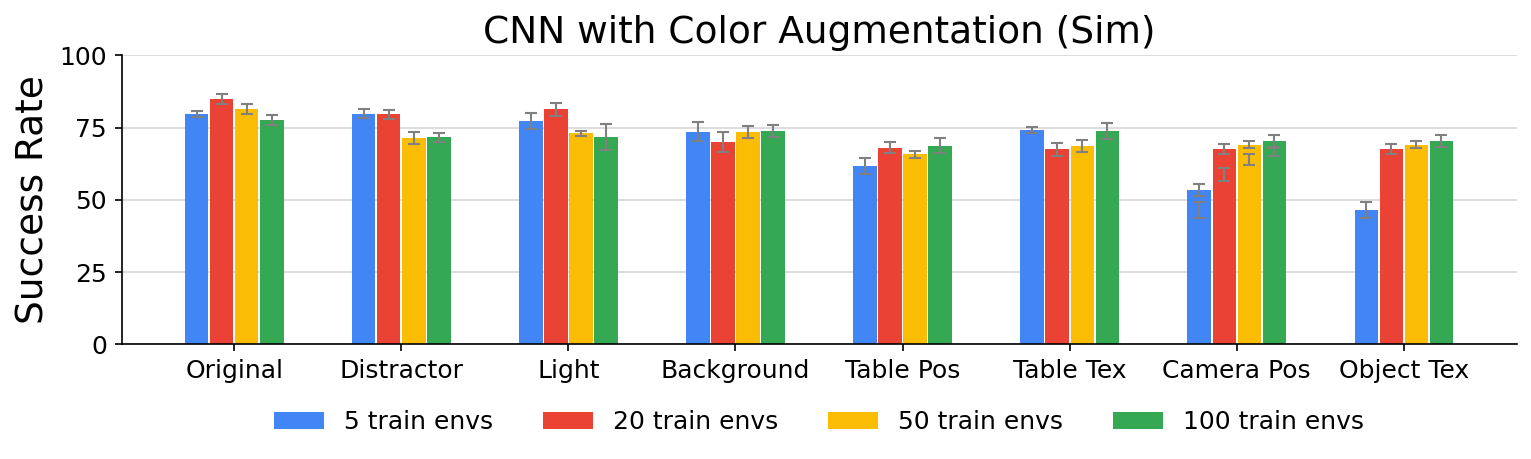}\\
    \includegraphics[width=0.9\textwidth]{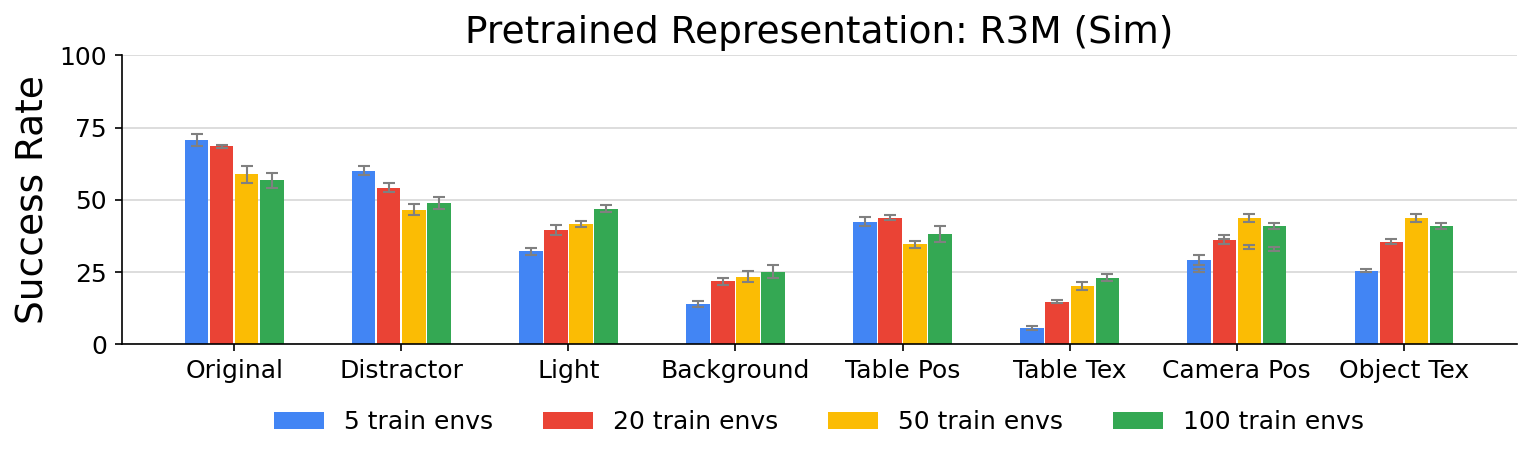}\\
    \includegraphics[width=0.9\textwidth]{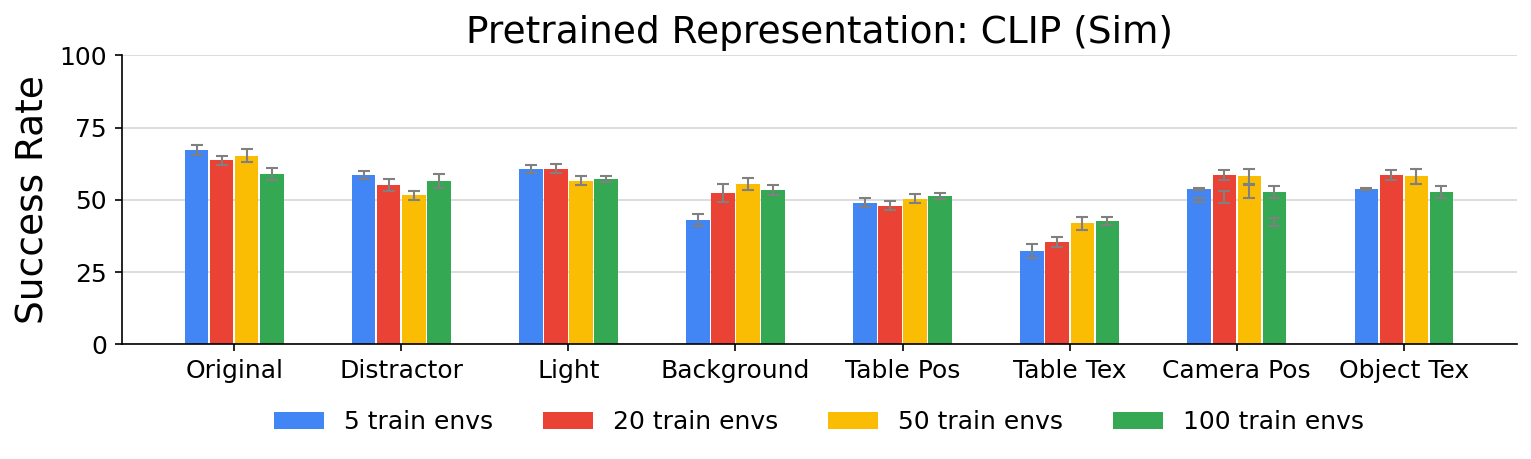}
    \caption{\small Success rates of simulated policies with data augmentations and with pretrained representations. Results are averaged over the 3 simulated tasks, with $5$ seeds run for each task.}
    \label{fig:sim_success_rate}
\end{figure*}

In Fig.~\ref{fig:sim_success_rate}, we report the performance of policies trained with data augmentations and with pretrained representations, in terms of raw success rates. We find that for some policies, the performance on the train environments (see ``Original'') degrades as we increase the number of training environments. Nonetheless,  as we increase the number of training environments, we see higher success rates on the factor-shifted environments. However, it may be possible to see even more improvements in the success rate with larger-capacity models that fit the training environments better.

\end{document}